\pdfoutput=1
\documentclass[11pt]{article}

\usepackage[letterpaper,margin=1in]{geometry}
\usepackage[utf8]{inputenc}
\usepackage[T1]{fontenc}
\usepackage{lmodern}
\usepackage{url}
\usepackage{booktabs}
\usepackage{amsfonts}
\usepackage{nicefrac}
\usepackage{microtype}
\usepackage{xcolor}
\newcount\Comments
\newif\ifarxiv
\arxivfalse

\usepackage{bbm}
\usepackage{amsmath}
\usepackage{amssymb}
\usepackage{algorithm}
\usepackage[framemethod=TikZ]{mdframed}
\usepackage{mathtools}
\usepackage{amsthm}
\usepackage{natbib}
\usepackage{letltxmacro}
\usepackage{authblk}
\usepackage{centernot}
\usepackage{enumitem}

\PassOptionsToPackage{algo2e,ruled, linesnumbered}{algorithm2e}
\RequirePackage{algorithm2e}

\LetLtxMacro\amsproof\proof
\LetLtxMacro\amsendproof\endproof

\usepackage{thmtools}
\usepackage{xcolor}
\AtBeginDocument{%
  \LetLtxMacro\proof\amsproof
  \LetLtxMacro\endproof\amsendproof
}

\newtheorem{theorem}{Theorem}[section]
\newtheorem{proposition}[theorem]{Proposition}
\newtheorem{definition}{Definition}[section]
\newtheorem{lemma}[theorem]{Lemma}

\newtheorem{corollary}[theorem]{Corollary}

\newtheorem{remark}{Remark}[section]
\newtheorem*{theorem*}{Theorem}
\newtheorem*{lemma*}{Lemma}

\usepackage{xpatch}
\xpatchcmd{\proof}{\itshape}{\normalfont\proofnamefont}{}{}

\usepackage[most]{tcolorbox}

\usepackage[colorlinks=true, allcolors=blue]{hyperref}

\usepackage{color}
\definecolor{darkgreen}{rgb}{0,0.5,0}
\definecolor{purple}{rgb}{1,0,1}

\newcount\Comments
\newcommand{\kibitz}[2]{\ifnum\Comments=1\textcolor{#1}{#2}\fi}

\newcommand{\Acal}{\mathcal{A}}
\newcommand{\Vcal}{\mathcal{V}}

\newcommand{\Dcal}{\mathcal{D}}
\newcommand{\Fcal}{\mathcal{F}}

\newcommand{\Hcal}{\mathcal{H}}

\newcommand{\Ncal}{\mathcal{N}}

\newcommand{\Xcal}{\mathcal{X}}
\newcommand{\Ycal}{\mathcal{Y}}
\newcommand{\Wcal}{\mathcal{W}}
\newcommand{\Rcal}{\mathcal{R}}

\newcommand{\Ecal}{\mathcal{E}}
\newcommand{\Pcal}{\mathcal{P}}
\newcommand{\Qcal}{\mathcal{Q}}

\newcommand{\Ex}{\mathop{\mathbb{E}}}
\renewcommand{\Pr}{\mathop{\mathbb{P}}}

\newcommand{\phat}{\hat{p}}

\newcommand{\dde}{\mathcal{G}}

\newcommand{\zo}{\{0,1\}}
\usepackage{ulem}
\definecolor{revnew}{rgb}{0.00,0.20,0.75}
\definecolor{revdel}{rgb}{0.75,0.05,0.05}
\definecolor{revnote}{rgb}{0.00,0.45,0.00}

\newif\ifRevHighlight
\RevHighlighttrue

\ifRevHighlight
  \newcommand{\new}[1]{{\color{revnew}#1}}
  \newcommand{\del}[1]{{\color{revdel}\sout{#1}}}
  \newcommand{\revnote}[1]{{\color{revnote}\small[\textsc{note}: #1]}}
\else
  \newcommand{\new}[1]{#1}
  \newcommand{\del}[1]{}
  \newcommand{\revnote}[1]{}
\fi

\providecommand{\Wcal}{\mathcal{W}}

\providecommand{\Bcal}{\mathcal{B}}
\renewcommand{\new}[1]{#1}

\arxivtrue

\title{A Unified Framework for Probability Estimation of Data-Dependent Events}

\author[1]{Vitaly Feldman}
\author[1]{Satyen Kale}
\author[2]{Vinod Raman\thanks{Work done when the author was an intern at Apple and a student at the University of Michigan, Ann Arbor.}}
\author[1]{Kunal Talwar}
\author[2]{Ambuj Tewari}
\affil[1]{Apple}
\affil[2]{University of Michigan, Ann Arbor}

\hypersetup{hypertexnames=false}

\begin{document}

\maketitle

\begin{abstract}
We study the general problem of estimating $D(\dde(S))$, the probability that an unknown distribution $D$ assigns to an event determined by a sample $S$ of $n$ i.i.d.\ draws from $D$ through a known mapping $\dde$. The estimation algorithm knows $\dde$ and $S$ but does not have access to any additional samples from $D$.  This formulation captures classical missing mass, the data-dependent targets of adaptive data analysis and post-selection inference, and the test error of a data-trained predictor, whose estimation was recently studied by \new{Gastpar et al.\ (2024a,b)}. Within this framework we examine the two classical approaches to estimation. The error of the empirical plug-in estimator is exactly a generalization gap, so generalization bounds transfer to it directly. The leave-one-out estimator, which recovers the Good--Turing estimator, is accurate for every mapping that is stable under the deletion of one sample point, including new structured examples such as neighborhoods of the sample and pierced sets.

The main application of our framework is to error estimation for a predictor that combines a base classifier with memorized exceptions, a structure common in theory and in practice. For it, the empirical estimator is identically zero and the leave-one-out estimator can fail. The test error is exactly the missing mass restricted to the error region of the base classifier, a region selected by the same sample. We formulate the general problem of \emph{adaptive missing mass estimation} to address such dependence and propose a natural hybrid estimator, the Good--Turing estimator restricted to the selected region. We show that the generalization of the base classifier alone does not guarantee the accuracy of this estimator, and derive two bounds on its error: the first one based on the VC dimension of possible error regions and another based on the mutual information of the selected region with the sample.
\end{abstract}

\section{Introduction}

Estimating the probability mass assigned by an unknown distribution to a data-dependent set is a basic problem that appears in several guises across probability, statistics, and learning theory. In supervised learning, for example, a training sample determines a classifier, and hence determines an \emph{error region} whose probability under the data distribution is the test error. In classical missing mass estimation, the sample determines the set of unseen outcomes, and the goal is to estimate the probability of drawing a new point from this unseen set. Although these examples are usually studied with different tools, they share a common structure: the target of estimation is not a fixed functional of the distribution alone, but a random quantity chosen as a function of the observed sample.

The classical theory has developed powerful estimators for several important special cases. Uniform-convergence and stability arguments justify the empirical plug-in estimator for many learning-theoretic error estimates. Good--Turing-style estimators, and their leave-one-out variants, estimate the missing mass and related $k$-occurrence or heavy-hitter functionals, even in regimes where the empirical plug-in estimator is identically zero or badly biased \citep{good1953population,mcallester2000convergence,mcallester2003concentration}. These successes raise a natural broader question:
\textit{When can one estimate the probability of an event that is itself chosen as a function of} the observed sample?

In this work, we study this question through a unified framework for \emph{data-dependent probability estimation}.
Given i.i.d.\ samples $S\sim D^n$ from an unknown distribution $D$ over a domain $\Xcal$, and a known mapping
$$
\dde:\Xcal^\star\to \zo^{\Xcal}
$$
that assigns to each sample an event, specified by its $\zo$-valued indicator function on $\Xcal$ (and, more generally, a $[0,1]$-valued query), the goal is to estimate the random quantity $D(\dde(S))$. The estimator is given $\dde$ and $S$, but not any additional fresh samples from $D$. Thus the same sample both determines the event and is used to estimate its probability (given fresh samples, the fraction of them that lies in $\dde(S)$ would be an accurate estimate). This formulation subsumes classical missing mass estimation, heavy-hitter estimation, and test-error estimation for learning algorithms.

For the error of a classifier produced by a known learning algorithm, this question was recently posed by \citet{gastpar2024fantastic,gastpar2024which}, who refer to the existence of an accurate estimator of the error as \emph{estimability} of the algorithm and use it to study when generalization bounds can be tight. The error region of a classifier is a data-dependent event and, conversely, every data-dependent event is the error region of a classifier over a suitably labeled domain, so the notion of learning algorithm estimability and our notion of event estimatability point to the same fundamental problem. In Section~\ref{sec:related-known} we give a detailed comparison and discuss some of the differences in the specific formulation we use.

Our guiding example and the main application of our framework is estimation of the error of a predictor that combines structured generalization with memorization. Such predictors are classical in learning theory, where memorizing exceptions is used to derive Occam (compressing) algorithms from PAC learning \citep{boardpitt90,blumer1987occam}, and common in practice, where deployed models are augmented with local patches, residual memories, and edit modules populated by specific errors \citep{kauschke2018batchwise,yang2023resmem,mitchell2022memory}. For such exception-memorizing predictors both classical approaches to error estimation fail. The empirical estimator is identically zero and therefore does not give useful information about the true error. The leave-one-out estimate, in turn, requires retraining the base learner $n$ times and comes with no guarantee when that learner is unstable. The quantity to be estimated nonetheless has a clean description: the true error is exactly the \emph{missing mass restricted to the error region of the base classifier}, the probability that a fresh point is unseen and falls into a region that was itself selected by the sample. To address this problem, in Section~\ref{sec:beyond}, we formulate this more generally as an \emph{adaptive missing mass estimation} task and give a hybrid estimator that combines the empirical and leave-one-out estimator. We show that the generalization of the base classifier alone does not guarantee the accuracy of the hybrid estimator, but prove guarantees for it when either the error region of the base classifier has limited mutual information with the sample or the class of possible error regions has low VC dimension.

Sample-dependent regions whose missing mass is of interest arise well beyond this example. A deployed language model may decline to answer prompts that fail a filter trained on the same data, so that hallucination in the sense of \citet{kalai2024calibrated}, the mass of facts unseen in training, must be measured on a region chosen by the data. An analyst may ask how many distinct values remain unseen in a subpopulation defined by a predicate chosen after inspecting the sample \citep{haas1995sampling,charikar2000towards}.

\subsection{Overview}
We summarize our contributions below.

\paragraph{A framework for data-dependent probability estimation.}
We define the estimatability of a mapping $\dde$ in terms of the worst-case mean squared error of an estimator of $D(\dde(S))$ (Definition~\ref{def:estmse}), and extend the definition to $[0,1]$-valued queries, to randomized mappings, and to restricted, possibly sample-size-dependent, classes of distributions (Definition~\ref{def:est-general}). In Section~\ref{sec:related-known} we show how this framework captures several known problems, including estimation of the test error or risk of a trained predictor, the estimability of learning algorithms, missing mass and heavy-hitter estimation, the Bayesian version of adaptive data analysis, point estimation of the targets of post-selection inference, and estimation of the realized miscoverage of conformal prediction sets. We discuss other related work in Section~\ref{sec:related}.

\paragraph{Empirical estimation.}
We observe that the error of the empirical estimator \emph{is} the generalization error of $\dde$ viewed as a learning algorithm. Therefore every generalization bound for $\dde$ (e.g.\ one based on the VC dimension, algorithmic stability, differential privacy, max-information, or mutual information) is a sufficient condition for estimatability (Section~\ref{sec:empest}).

\paragraph{Leave-one-out estimation.}
We observe that the classical hypothesis-stability bound for the leave-one-out estimate applies to any data-dependent event, so that the leave-one-out estimator has vanishing mean squared error for every mapping that is stable under the deletion of one sample point (Theorem~\ref{thm:looest}). Computing the stability of specific mappings recovers the classical missing mass and heavy-hitter guarantees, with the leave-one-out estimator coinciding with the Good--Turing estimator, and extends them to new structured examples (Section~\ref{sec:looest}). Specifically, for every neighborhood function on the domain, the mass of the union of the neighborhoods of the sample points is $O(1/n)$-estimatable, although this set can be much larger than the sample (Theorem~\ref{thm:neighbmass}). The same holds for the intersection of these neighborhoods, which includes the missing mass, and for the seen and the unseen parts of a fixed region. For the union of the neighborhoods of the points that occur exactly $k$ times in the sample, we obtain the rate \new{$O((k+\log n)/n)$, again for every neighborhood function} (Theorem~\ref{thm:heavy-hitters}).

\paragraph{Adaptive missing mass estimation.}
As the main application of our framework, in Section~\ref{sec:beyond} we formulate the problem of \emph{adaptive missing mass estimation} (Definition~\ref{def:adaptive-missing-mass}) and propose a hybrid estimator for it. In this problem a selection rule chooses a region of the domain using the sample (such as the error region of the base classifier), and the goal is to estimate the missing mass of the selected region. Our estimator is based on writing this missing mass as the difference between the mass of the selected region and the mass of its part that is seen in the sample. We estimate the first term empirically (for the error region of a base classifier this gives its training error) and the second by the leave-one-out estimator in which the region is held fixed at its value on the full sample. The difference of the two estimates is the fraction of the sample made up of points that lie in the selected region and occur in the sample exactly once. Thus our estimator is the Good--Turing estimator restricted to the selected region (Definition~\ref{def:hybrid}). For the error of an exception-memorizing classifier, computing it requires training the base classifier once rather than $n$ times.

For a region fixed in advance, the restricted Good--Turing estimator has mean squared error $O(1/n)$ (Corollary~\ref{thm:pierccomp}). However, here the region is selected using the same sample on which the estimator is evaluated. We show that this dependence can make the hybrid inaccurate even when the empirical mass of the selected region estimates its true mass with mean squared error $O(1/n)$ under every distribution (Proposition~\ref{prop:gap-not-enough}).

At the same time, we show that the hybrid is accurate whenever the selected region has limited mutual information with the sample or the class of possible regions has low VC dimension. Specifically, its mean squared error is $O((I+1)/n)$, where $I$ is the mutual information between the selected region and the sample, and $O(d\log(n/d)/n)$, where $d$ is the VC dimension of the class of possible regions (with no restriction on how the region is chosen from this class). See Theorem~\ref{thm:adaptive-missing-mass} for a formal statement.

To prove the bound in terms of the mutual information we first show that for a region fixed in advance the error of the restricted Good--Turing estimator is sub-Gaussian at scale $1/\sqrt n$. The restricted Good--Turing estimate concentrates by McDiarmid's inequality, and the missing mass of the region by an extension of the results of \citet{mcallester2003concentration}. We then use the Donsker--Varadhan variational inequality to decouple the selected region from the sample at the cost of an additive $O(I/n)$ term.
To prove the bound in terms of the VC dimension we replace the selected region by a nearby element of a suitable finite cover of the class of possible regions. We control the effect of the replacement using uniform convergence over the symmetric differences of possible regions and a union bound over the cover together which, again, relies on the concentration of the restricted Good-Turing estimator.

As a consequence, we get that the error of an exception-memorizing learner is estimatable whenever the base learner's hypothesis class has bounded VC dimension (Corollary~\ref{cor:vc-hypothesis-class}), even though the empirical estimator is identically zero. Specifically, if the base learner outputs binary classifiers from a class of VC dimension $d$, then the hybrid estimates the error of the memorizing learner with mean squared error $O(d\log(n/d)/n)$ under every distribution and every labeling of the data (which need not belong to the class). We also give an example of a learner for which the hybrid has error $O(1/n)$ while both standard estimators have constant error (Proposition~\ref{prop:hybrid-necessary}).

\paragraph{Limits of estimatability.}
In Section~\ref{sec:limits} we give three constructions of data-dependent events that cannot be consistently estimated by any procedure. The first of them is new and is a particularly simple instance of post-selection inference. Here the event consists of the points that are equal to $1$ at the coordinate with the largest empirical mean (among $2^n+1$ binary coordinates), and its mass is the true mean of the selected coordinate. We consider distributions in which one coordinate is always $1$ and all other coordinates are independent and uniform. With constant probability, exactly one other coordinate is then also equal to $1$ on all sample points. If the special coordinate is chosen uniformly at random, then given such a sample the two coordinates are equally likely to be the special one. The true mean of the selected coordinate is then equally likely to be $1$ and $1/2$. Thus no estimator achieves mean squared error below $1/64$ (Theorem~\ref{thm:max-selection-lb}).

The other two constructions are close to those of \citet{gastpar2024fantastic,gastpar2024which} and use the same approach on progressively smaller domains. The last of these has quadratic in $n$ size and also shows that \emph{finiteness} of the VC dimension of the image class does not suffice. Its image class has $\operatorname{VC}(\Hcal_n)\leq n+1$, whereas the empirical estimator is consistent as soon as $\operatorname{VC}(\Hcal_n)=o(n)$ (Corollary~\ref{cor:empvc}).

\section{Related Work}
\label{sec:related}
Estimating the test error of a data-trained predictor has a long history in statistical learning theory; see the textbooks \citep{shalev2014understanding,mohri2012foundations,anthony2009neural} for an overview. Several of the problems related to ours, namely prediction-error estimation, missing mass estimation, adaptive data analysis, post-selection inference, and conformal prediction, are easier to compare with our setting once our definitions are in place, and we describe how each of them maps to our framework in Section~\ref{sec:related-known}. In this section we discuss the lines of work most closely related to our results.

The closest line of work to ours is the study of \emph{estimability} by \citet{gastpar2024fantastic,gastpar2024which}. Motivated by the empirical finding that many of the generalization bounds proposed for neural networks are far from tight \citep{jiang2020fantastic}, \citet{gastpar2024fantastic} observe that a generalization bound that is tight for every distribution is an estimator of the test error, and define a hypothesis class or a learning algorithm to be estimable if such an estimator exists. They show that for a fixed, known algorithm there may exist a trade-off between the algorithm's ability to learn a class and the estimability of its error. Specifically, this trade-off applies to algorithms that learn parity functions and linear functions over finite fields. \citet{gastpar2024which} then study which algorithms are estimable. They give lower bounds for algorithms with an inductive bias towards a class of nearly orthogonal functions, or towards a suitable subset of a class of VC dimension $d$ when $n\leq\sqrt d/10$. They also show that an algorithm that is stable under the deletion of a block of $k$ sample points is estimable via a holdout estimator that trains on the remaining $n-k$ points. Their algorithm-dependent notion of estimability coincides with ours when specialized to the error regions of classifiers, up to the error criterion (Section~\ref{sec:related-known}). In contrast, the main focus of our work is the unified framework that naturally incorporates a variety of existing problems as well as new upper bounds that rely on tools developed for different problems.

The estimation of the error of a learning algorithm via leave-one-out and cross-validation has been analyzed through algorithmic stability since the work of \citet{rogers1978finite} and \citet{devroye1979distribution}. \citet{kearns1999algorithmic} give sanity-check bounds for the leave-one-out estimate under hypothesis stability, \citet{bousquet2002stability} relate several notions of stability to generalization and to the leave-one-out estimate, and \citet{kale2011cross} and \citet{kumar2013near} bound the error of $k$-fold cross-validation via mean-square and loss stability.

The study of missing mass estimation dates back to the work of \citet{good1953population} and more recently \citet{mcallester2000convergence}. Refinements of the missing-mass concentration bounds of \citet{mcallester2000convergence,mcallester2003concentration} appear in \citet{berend2012missing,berend2013concentration}, and \citet{maurer2022concentration} extends them to metric spaces. A parallel line of work investigates minimax rates for missing mass estimation. Near-tight upper and lower bounds were developed in \citep{rajaraman2017minimax,acharya2018improved}, alongside information-theoretic lower bounds based on classical techniques such as Fano's and Le Cam's methods \citep{paninski2003estimation,fano1966transmission,lecam1973convergence,yu1997assouad}. Complementary work by \citet{valiant2011estimating,valiant2017estimating} studies the estimation of the unseen portion of the distribution, leading to near-optimal estimators for entropy, support size, and other distributional properties. We refer the reader to  \citep{orlitsky2016optimal,balocchi2026bayesian} for a more comprehensive overview of works related to missing-mass, unseen-species, and support-size estimation.%

Our framework is motivated by and explores data-dependent estimation tasks that require combining distinct estimation principles. A number of existing works on generalization have relied on combinations of different approaches to generalization: \citet{foster2019hypothesis} combine hypothesis-set stability with complexity bounds for data-dependent hypothesis sets, while \citet{rivasplata2018pac}, \citet{dziugaite2018data}, and \citet{awasthi2020pac} develop PAC-Bayes guarantees for distribution- or sample-dependent priors using various notions of stability such as differential privacy.
See also  \citep{hellstrom2025generalization} for a broader survey of information-theoretic and PAC-Bayes approaches to generalization.

\section{Preliminaries}
\label{sec:prelim}

\subsection{Notation}
Let $\Xcal$ be a countable space, and let $\Delta \Xcal$ denote the set of all distributions over $\Xcal$. For a set $A \subseteq \Xcal$, we use $A^c$ to denote its complement and $|A|$ to denote its cardinality. For a distribution $D \in \Delta \Xcal$ and a subset $A \subseteq \Xcal$, we use $D(A)$ to denote the cumulative mass that $D$ places on $A$. That is, $D(A) := \Pr_{x \sim D} \left[x \in A\right]$. We also define sets using their indicator functions and more generally consider real-valued functions on $\Xcal$. For a function $h:\Xcal \to \mathbb{R}$, we define $D(h)= \Ex_{x \sim D} [h(x)]$. For a single point $x$ we abbreviate $D(\{x\})$ by $D(x)$.

As usual, for $n \in \mathbb{N}$, we define $[n] := \{1, \dots, n\}.$ We define $\Xcal^{\star}$ as the set of all finite sequences of examples in $\Xcal$. We denote a sample of $n$ examples $x_1, \dots, x_n \in \Xcal$ by $S$, and, abusing notation, also write $S$ or $T(S)$ for the set $\{x_1, \dots, x_n\}$ of its points (so that $S^c$, $S \cap A$, and $|S|$ refer to that set). A mapping $\dde$ defined on $\Xcal^\star$ is \emph{symmetric} if $\dde(S)=\dde(\pi S)$ for every sample $S$ and every permutation $\pi$ of its coordinates, i.e.\ if $\dde(S)$ depends on $S$ only through its multiset of points.  \new{Unless stated otherwise, the mappings considered in this paper are symmetric.} For a sample $S \sim D^n$ and index $i \in [n]$, we write $S^{-i}$ for the sample $x_1, \dots, x_{i-1}, x_{i+1}, \ldots, x_n$ with the $i$th point removed. In particular, $S^{-n}$ denotes the sample with its last point removed. The sample size $n$ is always clear from context.

\subsection{Probability Estimation}
In probability estimation, one aims to estimate the mass placed by an \emph{unknown} distribution $D \in \Delta \Xcal$ on a \emph{known}, potentially sample-dependent, region using i.i.d.\ samples from $D$. Throughout the paper, we use $\dde: \Xcal^{\star}  \rightarrow \zo^{\Xcal}$ to denote a known mapping that assigns to each sample $S$ an indicator function on $\Xcal$. We write $f := \dde(S)$ for the resulting query and $f(x)$ for its value at a point $x \in \Xcal$. We identify the indicator $f$ with the event (subset) $\{x \in \Xcal : f(x) = 1\}$ it defines, and freely use set operations on the outputs of $\dde$ whenever convenient. Both $\Xcal$ and $\dde$ are assumed known, and the target of estimation is the induced mass
$$
D(\dde(S)) = D(f) = \Ex_{x \sim D}\left[f(x)\right],
$$
which for an event equals $\Pr_{x \sim D}\left[x \in \dde(S)\right]$.
More generally, we also allow $\dde$ to output a function $f = \dde(S)$ from $\Xcal$ to $[0,1]$, i.e.\ a \emph{statistical query} or a (bounded) loss function.

We first define a probability estimator for the simpler case of deterministic mappings $\dde$. A probability estimator $\phat$, defined formally in Definition~\ref{def:est}, maps a sample to a value in $[0, 1].$ In this paper, we will always assume $\phat$ is deterministic, though all our definitions of estimatability can be naturally extended to randomized estimators. Note that replacing a randomized $\phat$ by its expectation over its internal randomness can only decrease the mean squared error.

\begin{definition} [Probability Estimator] \label{def:est} A probability estimator $\phat: \Xcal^{\star} \rightarrow [0, 1]$ is a function that maps a finite training sample $S \in \Xcal^{\star}$ to a number in $[0, 1]$.
\end{definition}

Given knowledge of $\dde$ (but not $D$), our goal is to find an estimator $\phat$ such that $\phat(S)$ and $D(\dde(S))$ are close when $S \sim D^n$ is an i.i.d.\ draw. We will focus on the mean-squared error (MSE) of the estimator both for concreteness and because it is the most common notion of error in the literature.
\begin{definition}[Estimatability] \label{def:estmse} For an integer $n$ and $\alpha \in [0,1]$, a mapping $\dde: \Xcal^{\star} \rightarrow \{0,1\}^{\Xcal}$ is $\alpha$-\emph{estimatable with $n$ samples} if there exists an estimator $\phat$ such that
$$\sup_{D \in \Delta \Xcal} \Ex_{S \sim D^n}\left[(\phat(S) - D(\dde(S)))^2 \right] \leq \alpha \ .$$
\end{definition}
When the sample size is clear from the context we simply say that $\dde$ is $\alpha$-estimatable, and we write that $\dde$ is $\alpha(n)$-estimatable for a function $\alpha(\cdot)$ if it is $\alpha(n)$-estimatable with $n$ samples for every $n$. We also allow the problem itself to be parameterized by $n$, that is $\dde = \{\dde_n\}_{n=1}^{\infty}$ where $\dde_n: \Xcal_n^{\star} \rightarrow \zo^{\Xcal_n}$ is defined over a domain $\Xcal_n$ that may depend on $n$. We say that $\dde$ is \emph{estimatable} if $\dde_n$ is $\alpha_n$-estimatable with $n$ samples for some sequence $\alpha_n \to 0$ as $n \to \infty$, and \emph{non-estimatable} otherwise.
This definition can be easily extended to other notions of error, such as mean absolute deviation (MAD) or a high-probability tail bound.

In many specific learning problems one is only interested in the performance of the estimator over a restricted class of input distributions $\Dcal$ (as opposed to all distributions over the domain). The definition of estimation error and estimatability of mappings naturally extends to this class-specific setting by taking the supremum over $D\in \Dcal$. In addition, in many practical applications the event or query is generated by a randomized algorithm (such as a randomly initialized learning algorithm). In this case, we can extend the definition of estimatability to randomized mappings $\dde$ by allowing the estimator to depend on the randomness of $\dde$ as well. We incorporate these generalizations into the definition below.

\begin{definition} [General Estimatability for Randomized Mappings] \label{def:est-general} Let $\Dcal = (\Dcal_n)_{n\in\mathbb{N}}$ be a family of classes of distributions over $\Xcal$, one for each sample size.
    A randomized mapping is defined by a function $\dde: \Xcal^{\star} \times \Omega \rightarrow [0,1]^{\Xcal}$, where $\Omega$ is the space of random seeds together with a distribution $R$ over $\Omega$. For an integer $n$ and $\alpha \in [0,1]$, the mapping $\dde$ is then said to be $\alpha$-estimatable over $\Dcal$ with $n$ samples if there exists an estimator $\phat: \Xcal^{\star} \times \Omega \rightarrow [0,1]$ such that
    $$\sup_{D \in \Dcal_n} \Ex_{S \sim D^n, r\sim R}\left[(\phat(S,r) - D(\dde(S,r)))^2 \right] \leq \alpha \ .$$
\end{definition}

In this paper, we are interested in understanding general conditions under which functions $\dde: \Xcal^{\star} \rightarrow \zo^{\Xcal}$ are estimatable. We now discuss several known choices of $\dde$ arising in specific problems and their estimatability properties.

\subsection{Relationship to Known Estimation Problems}\label{sec:related-known}

\paragraph{Prediction-error and loss estimation.}
Estimating the error of a data-trained predictor is a prototypical instance of our question. If a learning algorithm maps a labeled sample $S$ to a classifier $h_S$, then $\dde(S)=\{(x,y):h_S(x)\neq y\}$ is the (data-dependent) error region and $D(\dde(S))$ is its test error; for a bounded loss, taking $\dde(S)$ to be the corresponding loss function makes $D(\dde(S))$ the risk. Classical generalization theory estimates this quantity by the training error whenever the induced class of error regions has low complexity, e.g.\ bounded VC dimension \citep{vapnik1971uniform} or bounded Rademacher complexity \citep{boucheron2013concentration}.
In some regimes where the empirical estimator fails, cross-validation, leave-one-out, and bootstrap methods have been shown to estimate risk \citep{stone1974cross,devroye1979distribution,kearns1999algorithmic,efron1983estimating}.

\paragraph{Estimability of learning algorithms.}
The question of when the test error of the classifier produced by a known learning algorithm can be estimated from the training sample alone was posed by \citet{gastpar2024fantastic} and studied further by \citet{gastpar2024which}. In their terminology, a learning algorithm $\Acal$ is $(\varepsilon,\delta,n)$-estimable with respect to a class of distributions if there is an estimator $E$ (which may depend on $\Acal$) such that $|E(S)-D(\dde(S))|\leq\varepsilon$ with probability at least $1-\delta$ over $S\sim D^n$ for every distribution $D$ in the class, where $\dde(S)$ is the error region of $\Acal(S)$ (or, for a bounded loss, its loss function). \citet{gastpar2024which} also consider the version of this guarantee that holds on average over a uniformly random distribution from a finite collection, and its mean-squared-error analogue. Since $\Acal$ is known, estimability with respect to all distributions is estimatability of the mapping $\dde$ in the sense of Definition~\ref{def:estmse}, up to the choice of the error criterion. Conversely, for any mapping $\dde$ over $\Xcal$, consider the labeled domain $\Xcal\times\zo$ with distributions supported on the label $0$, and the learning algorithm that on a sample $S$ outputs the classifier $x\mapsto\dde(S)(x)$; its error region is $\dde(S)\times\{0\}$, so estimating its error is the same problem as estimating $D(\dde(S))$ (for a $[0,1]$-valued query the same identification goes through the loss $\ell(q,x)=q(x)$ of Section~\ref{sec:empest}). Thus the two frameworks describe the same class of problems, and we keep the term \emph{estimatability} for Definition~\ref{def:estmse}, which applies to mappings and uses the mean squared error. The formulation in terms of events makes it natural to consider missing mass and its structured variants alongside prediction error, and to combine the estimators for them, which is the focus of this work.

\paragraph{Missing mass estimation.}
The most classical special case is \emph{missing mass}: taking $\dde(S)=S^c$, the target $D(\dde(S))$ is the probability that a fresh sample is not present in $S$. Here the empirical estimator is identically zero, yet the Good--Turing estimator and its refinements estimate the missing mass at mean-squared-error rate $\Theta(1/n)$: the Good--Turing bias is $O(1/n)$ \citep{good1953population}, with high-probability guarantees given by \citet{mcallester2000convergence}, and the minimax mean-squared error is $\Theta(1/n)$ \citep{rajaraman2017minimax,acharya2018improved}. More generally, the mass of the $k$-occurrence (heavy-hitter) set $\dde(S)=\{x \in \Xcal:\sum_{i=1}^n\mathbf 1\{x_i=x\}=k\}$ is $O(k/n)$-estimatable \citep{mcallester2000convergence}. The leave-one-out estimator of Section~\ref{sec:looest} is based on these results and extends them to several additional structures.

\paragraph{Adaptive data analysis.}
In adaptive data analysis (ADA), a curator holding a sample $S\sim D^n$ answers a sequence of statistical queries $h:\Xcal\to[0,1]$ posed adaptively by an analyst, returning estimates of each value $D(h)=\Ex_{x\sim D}[h(x)]$ \citep{DworkFHPRR14:arxiv,DworkFHPRR15:arxiv,DworkFHPRR15:science,BassilyNSSSU16,russo2016controlling,feldman2018calibrating}. Our target $D(\dde(S))$ is the value of a single such query, $h=\dde(S)$. In the positive direction, the machinery developed for ADA transfers to our setting directly. That machinery controls the \emph{generalization gap} $\Ex_S\bigl|\tfrac1n\sum_{i=1}^n \dde(S)(x_i)-D(\dde(S))\bigr|$ between the empirical answer and its true value by limiting how strongly the query $\dde(S)$ depends on the sample $S$. The main techniques are:
\begin{itemize}
  \item \textbf{Differential privacy.} If $\dde(\cdot)$ is $(\varepsilon,\delta)$-differentially private as a function of $S$, its output is insensitive to any single sample point, and the generalization gap is controlled by $\varepsilon$ and $\delta$ \citep{DworkFHPRR14:arxiv}; tight and variance-adaptive versions of this transfer theorem were given by \citet{BassilyNSSSU16,feldman2018calibrating}.
  \item \textbf{Max-information.} More generally, a bound on the (approximate) max-information $I_\infty^\beta$ between the sample $S$ and the query $\dde(S)$ controls the gap \citep{DworkFHPRR15:arxiv,rogers2016max}.
  \item \textbf{Mutual information.} A bound on the mutual information $I(S;\dde(S))$ between the sample and the query likewise controls the gap: the expected \emph{signed} gap is at most $\sqrt{I(S;\dde(S))/(2n)}$ \citep{russo2016controlling,xu2017information,feldman2018calibrating}.
\end{itemize}
Each of these is a sufficient condition for the empirical estimator to accurately estimate $D(\dde(S))$.

At the same time the requirements in ADA are incomparable to those of our setting. Adaptive data analysis requires a guarantee that holds for \emph{every} distribution $D$ and \emph{every} analyst; equivalently, the analyst may be assumed to know $D$ but not the realized sample $S$, so the query it produces is independent of $S$ (it can depend on $S$ only indirectly, through the curator's earlier answers). However in estimatability, the mapping $\dde$ is fixed and known and its output $\dde(S)$ depends on $S$ \emph{directly}, and the estimator is given both $\dde$ and $S$. Thus the basic ADA setting does not itself fall under our notion of estimatability. In particular, while positive results transfer, \emph{hardness does not}: a lower bound for adaptive data analysis \citep{hardt2014preventing,steinke2015interactive} does not imply a lower bound for estimatability, which is why the impossibility results of Section~\ref{sec:limits} require constructions of their own rather than being inherited from ADA.

\paragraph{Bayesian adaptive data analysis.}
\citet{Elder2016Challenges} proposed a Bayesian version of ADA in which the analyst and the curator both receive a prior over distributions on $\Xcal$, and the data distribution $D$ is drawn from this prior (and is unknown to both). The curator wins if, with probability at least $1-\delta$, all of its answers are within $\varepsilon$ of the values of the queries, where the probability is also over the draw of $D$ \citep[Section~2]{Elder2016Challenges}. \citet{NissimStemmerTsfadia2023} study a balanced model in which the distribution is chosen by a sampler that passes no information to the analyst. They note that when the sampler is publicly known, the two models are equivalent from an information-theoretic point of view. By a standard minimax argument, a curator that wins against every analyst exists for every prior if and only if for every analyst there is a curator that wins against it for every distribution $D$ (the curator may depend on the strategy of the analyst but not on its random bits). Indeed, for a fixed prior the minimax theorem allows the curator to be chosen after the analyst, and for a fixed analyst it equates the worst case over priors with the worst case over distributions (for a finite domain and finitely many possible queries and answers). We are not aware of this equivalence being stated in the literature. For a known analyst and \new{fixed answering rules of the curator for the first $j-1$ queries}, the $j$-th query is a known randomized function of the sample whose value the curator needs to estimate for every distribution.\footnote{Strictly speaking, Definition~\ref{def:est-general} gives the estimator the random bits of the analyst, which the curator does not see. This does not change the minimax error since, for every prior, the conditional expectation of the value of the query given the sample and the random bits depends only on the sample and the query.} Thus our framework (Definition~\ref{def:est-general}) captures this setting, one query at a time. For Dirichlet priors, \citet{Elder2016Subgaussianity} shows that the curator that answers with the posterior mean matches the statistical guarantees of the static case. Beyond such specific priors, we are not aware of positive results for this setting that are stronger than those for general ADA. The negative results are either purely computational \citep{NissimStemmerTsfadia2023} or apply only to restricted families of curators, such as those that answer each query with a function of its prior and posterior means \citep{Elder2016Challenges}.

\paragraph{Post-selection and selective inference.}
In post-selection (selective) inference, a model, variable, or hypothesis is chosen after examining the data, for example the support recovered by the Lasso, or the most significant among many candidates, and one seeks inference for a target \emph{defined by} that selection \citep{berk2013valid,taylor2015statistical,lee2016exact,fithian2014optimal}. This setting differs from ordinary generalization in two ways. First, the selection is typically from a rich family (all sub-models, all subsets of variables), so uniform convergence over the family is unavailable and the naive in-sample value of the selected target is systematically biased. Second, the standard remedy is not a complexity bound but \emph{conditioning on the selection event} and performing inference under the resulting conditional law \citep{lee2016exact,fithian2014optimal}. In our notation the selected object is $\dde(S)$ and the selected parameter is its mass $D(\dde(S))$. We differ in two ways: we seek distribution-free \emph{point estimation} of $D(\dde(S))$ rather than conditional tests or intervals, and we ask when this is possible \emph{at all} -- for a sufficiently rich selection family no estimator can succeed, as our impossibility results show (e.g.\ Theorem~\ref{thm:max-selection-lb}, which selects the coordinate of maximal empirical mean among exponentially many candidates).
\paragraph{Conformal prediction and distribution-free uncertainty quantification.}
Given a training or calibration sample $S$, a conformal method outputs a prediction set $C_S(x)$ for the label of a new point $x$; its error event $\dde(S)=\{(x,y):y\notin C_S(x)\}$ is data-dependent, and $D(\dde(S))$ is the realized miscoverage of the particular set produced from $S$ \citep{vovk2005algorithmic,lei2013distribution,angelopoulos2021gentle}. Conformal methods are designed to control this probability marginally over the joint draw of $S$ and the test point, under exchangeability. Our question is complementary: once a particular $C_S$ has been produced, can its realized miscoverage $D(\dde(S))$ be estimated from the same sample $S$? \citet{noorani2025conformal} use ideas from missing mass estimation to address this question for generative models.
\paragraph{Classical estimation.}
Finally, our task differs from classical parameter estimation in that the estimand is itself \emph{random}: $D(\dde(S))$ is a function of the realized sample, whereas in classical problems, mean estimation, or empirical risk minimization over $U$-statistics \citep{clemencon2008ranking}, the target is a fixed functional of $D$. This randomness of the estimand drives several of the phenomena we study and means that standard lower-bound techniques for classical estimation, such as Le~Cam's two-point method, do not directly apply: Appendix~\ref{app:classnequnclass} exhibits a mapping that is estimatable with zero error in the sense of Definition~\ref{def:estmse} while the corresponding deterministic functional $\Ex_{S\sim D^n}[D(\dde(S))]$ cannot be estimated to within a constant.

In Sections~\ref{sec:empest} and \ref{sec:looest}, we focus on two natural estimators $\phat$: the empirical-distribution-based estimator and the leave-one-out estimator. We show that these two estimators are sufficient for estimating a large class of functions $\dde$. In Section~\ref{sec:beyond} we combine the two estimators to solve adaptive missing mass estimation, and in Section~\ref{sec:limits} we complement our upper bounds with lower bounds.

\section{Empirical Estimator} \label{sec:empest}
The simplest and most well-studied estimator $\phat$ constructs the empirical distribution $\hat{D}$ over its samples $x_1, \dots, x_n$ and returns the empirical mass $\hat{D}(\dde(S))$ of the query $f = \dde(S)$ (for an event, $\Pr_{\hat{x} \sim \hat{D}}\left[\hat{x} \in \dde(S) \right]$), or equivalently,
$$
\phat_\dde^{\text{emp}}(S) := \frac{1}{n}\sum_{i=1}^n f(x_i).
$$
We can view the output $\dde(S)$ directly as a (data-dependent) loss function on a data point, in which case the error of the empirical estimator becomes the gap between the empirical and population loss. More formally,
define $\Qcal = [0,1]^{\Xcal}$ (which contains $\zo^{\Xcal}$) and an algorithm $\Acal:  \Xcal^{\star} \rightarrow \Qcal$ such that on input $S \in \Xcal^n$, we have $\Acal(S) = \dde(S).$ Define the loss function $\ell: \Qcal \times \Xcal \rightarrow [0,1]$ by $\ell(q, x) := q(x)$ (for an event $q$, this is $\mathbf{1}\{x \in q\}$). Then the minimax MSE of the empirical-distribution-based estimator can be written as
\begin{align*}
&\sup_{D \in \Delta \Xcal}\Ex_{S \sim D^n}\left[\left(\phat_\dde^{\text{emp}}(S)- D(\dde(S))  \right)^2\right] \\
&= \sup_{D \in \Delta \Xcal}\Ex_{S \sim D^n}\left[\left(\frac{1}{n}\sum_{i=1}^n \ell(\Acal(S), x_i) - \Ex_{x \sim D}\left[\ell(\Acal(S), x)\right]  \right)^2\right]
\end{align*}
where the last line is exactly the generalization error for algorithm $\Acal$ with respect to the loss function $\ell$.
This relationship implies that standard bounds for the empirical estimation of losses imply bounds for the corresponding events.

For a sample size $n \in \mathbb{N}$, define $\Hcal_n := \{\dde(S) \mid S \in \Xcal^n\}$ as the hypothesis class induced by the image space of $\dde$ on $\Xcal^n.$
Classical VC theory then yields the following bound, proved in Appendix~\ref{app:defvc}; in particular, any $\dde$ with $\operatorname{VC}(\Hcal_n) = o(n)$ is estimatable via the empirical estimator.

\begin{corollary}[VC-based upper bound for the empirical estimator] \label{cor:empvc}
Let $\dde: \Xcal^{\star} \rightarrow \zo^{\Xcal}$ be any function such that $\operatorname{VC}(\Hcal_n) < \infty$ for all $n \in \mathbb{N}$. Then
$$\sup_{D \in \Delta \Xcal}\Ex_{S \sim D^n}\left[\left(\phat_\dde^{\emph{\text{emp}}}(S)- D(\dde(S))  \right)^2\right] = O\left(\frac{\operatorname{VC}(\Hcal_n)+1}{n} \right).$$
\end{corollary}

A second, incomparable, sufficient condition is that the output of $\dde$ be small on average. Writing
\begin{equation} \label{eq:small}
\tau_\dde(n) := \sup_{D \in \Delta \Xcal} \mathop{\Ex}_{S \sim D^n} |\dde(S)|
\end{equation}
for the worst-case expected cardinality of the output of $\dde$, we show in Appendix~\ref{app:defvc} that \new{the mean-squared error of the empirical estimator is $O\big((\tau_\dde(n)/n)\log(en/\tau_\dde(n))\big)$}, so that estimation is possible whenever $\tau_\dde(n) = o(n)$.

\begin{theorem}[Cardinality-based upper bound for the empirical estimator] \label{thm:smallcard-mse}\label{thm:smallcard}
\new{Let $\dde: \Xcal^{\star} \rightarrow \zo^{\Xcal}$ and $n\geq1$ be such that $0<\tau_\dde(n)\leq n$. Then
$$\sup_{D \in \Delta \Xcal}\Ex_{S \sim D^n}\left[\Bigl(\phat_\dde^{\emph{\text{emp}}}(S)- D(\dde(S))  \Bigr)^2\right] \leq \frac{\tau_\dde(n)}{n}\left(2\ln\frac{n}{\tau_\dde(n)}+3\right).$$
If moreover $|\dde(S)|\leq\tau_0$ for every $S\in\Xcal^n$, then the left-hand side is at most $\tau_0/n$.}
\end{theorem}

The two complexity measures $\operatorname{VC}(\Hcal_n)$ and $\tau_\dde(n)$ are incomparable: Appendix~\ref{app:defvc} exhibits mappings for which each of the two bounds is vacuous while the other is not.

Another class of approaches for understanding the properties of the empirical estimator is based on various notions of stability of the output \citep{bousquet2002stability, shalev2010learnability, feldman2018generalization, feldman2019high}, all of which become properties of $\dde$ since $\Acal(S)= \dde(S)$. Information-theoretic generalization bounds, controlling the gap through the mutual or conditional mutual information between the sample and the output \citep{xu2017information,steinke2020reasoning,hellstrom2025generalization}, apply in the same way. Since these arguments mostly require only plugging in existing upper bounds on generalization error, we omit them here.

Finally, the adaptive-data-analysis techniques discussed in Section~\ref{sec:related-known}, namely differential privacy, max-information, and mutual information, each yield a further sufficient condition for the empirical estimator: whenever $\dde(S)$ has a small value of the corresponding quantity, the generalization gap above is controlled \citep{DworkFHPRR14:arxiv,DworkFHPRR15:arxiv,BassilyNSSSU16,russo2016controlling,feldman2018calibrating,rogers2016max}.

\ifarxiv
All of the above guarantees extend directly to \emph{randomized} mappings (Definition~\ref{def:est-general}): given the seed $r$, the empirical estimator uses the realized query $\dde(S,r)$, and identifying the algorithm $\Acal(S,r)=\dde(S,r)$ absorbs the internal randomness of $\dde$ into that of $\Acal$. Since the VC-, stability-, privacy-, and information-based bounds above all apply to randomized algorithms, they carry over without change.
\fi

\section{Estimation via the Leave-One-Out Estimator} \label{sec:looest}
In Section~\ref{sec:empest}, we identified generic conditions on $\dde$ for which the empirical-distribution-based estimator enjoys vanishing estimation error. One might wonder if this empirical distribution-based estimator can work for all $\dde$. Unfortunately, this is not the case even for the well-studied missing mass estimation problem $\dde(S) = S^c$. In this setting, the empirical plug-in estimator is identically zero, while the true probability mass of unseen elements is typically nonzero, leading to a large bias. Motivated by this example, we study the leave-one-out estimator $\phat_\dde^{\text{loo}}$, which, writing $f_i := \dde(S^{-i})$ for the query produced on the sample with the $i$th point removed, computes
$$
\phat_\dde^{\text{loo}}(S) = \frac{1}{n}\sum_{i=1}^n f_i(x_i).
$$
It turns out that while the missing mass $\dde$ is not estimatable via the empirical-distribution-based estimator, it is estimatable using the leave-one-out estimator. In fact, we show that the leave-one-out estimator is valid for \emph{any} sufficiently stable function $\dde$. Our notion of stability involves the following quantity:
\begin{equation} \label{eq:stab}
\gamma_\dde(n) := \sup_{D \in \Delta \Xcal} \gamma_\dde^D(n),
\qquad\text{where}\qquad
\gamma_\dde^D(n) := \Ex_{S \sim D^n}\left[D\big(|f - f'|\big)  \right],
\end{equation}
where $f = \dde(S)$, $f' = \dde(S^{-n})$, and $D\big(|f - f'|\big) = \Ex_{x \sim D}\,|f(x) - f'(x)|$ is the expected change in the query when the last sample point is dropped; for an event (indicator-valued $\dde$) this equals the symmetric-difference mass $D(\dde(S) \,\Delta\, \dde(S^{-n}))$.
We call $\gamma_\dde(n)$ the \emph{deletion stability} of $\dde$ at sample size $n$, and say that $\dde$ is \emph{stable} if $\gamma_\dde(n) = o(1)$.  This quantity is the \emph{hypothesis stability} of \citet{devroye1979distribution}, named by \citet{kearns1999algorithmic} and used in the form of \citet[Definition~3]{bousquet2002stability}, for the ``algorithm'' $S\mapsto\dde(S)$ and the loss $\ell(\dde(S),x)=\dde(S)(x)$. Theorem~\ref{thm:looest} below is the classical bound on the deleted estimate for hypothesis-stable algorithms (\citealp[Theorem~3.1]{kearns1999algorithmic}; \citealp[Lemma~9]{bousquet2002stability}), restated in this setting: the quantity those results bound, the difference between the true error of the hypothesis trained on the full sample and its leave-one-out estimate, is exactly the estimation error of $\phat_\dde^{\text{loo}}$ for $D(\dde(S))$. What is new is the object to which the bound is applied. A data-dependent event is the output of a learning algorithm in this sense, with no labels and no hypothesis class, and the content of this section lies in computing $\gamma_\dde$ for the mappings considered below and in identifying the resulting estimators with the Good--Turing estimator (Remark~\ref{rem:good-turing}).

Theorem~\ref{thm:looest} bounds the error of the leave-one-out estimator by $\gamma_\dde(n)$ up to an additive $O(1/n)$; in particular, every stable mapping is estimatable via the leave-one-out estimator. Missing proofs for this section are in Appendix~\ref{app:missproof}.

\begin{theorem}  [Estimation via the Leave-One-Out Estimator] \label{thm:looest} Let $\dde: \Xcal^{\star} \rightarrow [0,1]^{\Xcal}$ be any \emph{symmetric} mapping. Then, for every $n \geq 1$ and every $D \in \Delta\Xcal$,
$$
\Ex_{S \sim D^n}\left[\left(\phat_\dde^{\emph{\text{loo}}}(S) - D(\dde(S))\right)^2 \right] \leq \frac{1}{2n} + 3\gamma_\dde^D(n),
$$
and consequently
$$
\sup_{D \in \Delta \Xcal}\Ex_{S \sim D^n}\left[\left(\phat_\dde^{\emph{\text{loo}}}(S) - D(\dde(S))\right)^2 \right] \leq \frac{1}{2n} + 3\gamma_\dde(n).
$$
In particular, if $\gamma_\dde(n) = o(1)$ then $\dde$ is estimatable.

\end{theorem}

While missing mass estimation gives a task for which leave-one-out succeeds and the empirical estimator fails, the following proposition gives the reverse direction. Together, these examples show that the two estimators are incomparable.

\begin{proposition}[Empirical succeeds while leave-one-out fails] \label{prop:incomparable}
 For any non-empty domain $\Xcal$, define $\dde(S) = \Xcal$ if the length of the \emph{sequence} $S$ is even, and $\dde(S) = \emptyset$ if the length is odd.
  Then $\phat_\dde^{\emph{\text{emp}}}(S) = D(\dde(S))$ while $\phat_\dde^{\emph{\text{loo}}}(S) = 1 - D(\dde(S))$, for every $S$ and $D$.

\end{proposition}
\ifarxiv
\begin{proof}
If $n$ is even then $\dde(S)=\Xcal$, so $D(\dde(S))=1=\phat_\dde^{\text{emp}}(S)$, while every fold $S^{-i}$ has odd length, so $\dde(S^{-i})=\emptyset$ and $\phat_\dde^{\text{loo}}(S)=0$. If $n$ is odd the roles are reversed: $D(\dde(S))=0=\phat_\dde^{\text{emp}}(S)$ and $\phat_\dde^{\text{loo}}(S)=1$.
\end{proof}

\fi%
Appendix~\ref{sec:hybrid-separation} shows a simple problem that combines the failure modes of both the empirical and the leave-one-out estimators (but is solved by the proposed hybrid estimator).

As with the empirical estimator, the leave-one-out estimator extends directly to \emph{randomized} mappings (Definition~\ref{def:est-general}): it reuses a single seed $r$ across all folds, computing $\dde(S^{-i},r)$, so that conditioning on $r$ reduces to the deterministic guarantee of Theorem~\ref{thm:looest}. Averaging over $r\sim R$ then preserves the bound, with $\gamma_\dde(n)$ now also averaged over $r$. We omit the formal details.

\begin{remark}[The leave-one-out estimator is the Good--Turing estimator] \label{rem:good-turing}
For the two classical mappings the leave-one-out estimator coincides exactly with the Good--Turing estimator. Write $F_j(S)$ for the number of distinct elements occurring exactly $j$ times in $S$. For missing mass, $\dde(S) = S^c$, a point $x_i$ lies in $(S^{-i})^c$ precisely when it occurs once in $S$, so
$$\phat_\dde^{\text{loo}}(S) = \frac{F_1(S)}{n},$$
the Good--Turing estimator of \citet{good1953population}. More generally, for $\dde(S) = c_k(S)$, the set of points occurring exactly $k$ times in $S$ (Section~\ref{sec:heavy}), a point $x_i$ lies in $c_k(S^{-i})$ precisely when it occurs $k+1$ times in $S$, and each such element is counted once for each of its $k+1$ occurrences, so
$$\phat_\dde^{\text{loo}}(S) = \frac{(k+1)F_{k+1}(S)}{n},$$
which is the classical Good--Turing estimator for the mass of the $k$-occurrence set. Theorem~\ref{thm:looest} together with Lemma~\ref{lem:khh} thus recovers the classical guarantees for these estimators, and simultaneously extends the same estimator to the mappings considered below.
\end{remark}

\subsection{Applications}

In this section, we upper bound $\gamma_\dde(n)$ for several natural choices of $\dde$. Our results recover well-known facts, such as missing mass being $O(1/n)$-estimatable, while also showing that estimation is possible for new choices of $\dde$.

\paragraph{Estimating Mass of Neighbors.}
\label{sec:neighbors}
In this section, we study the estimatability of natural $\dde$ functions whose output size can be much larger than $n$. These functions $\dde$ are best described by a directed graph on the vertex set $\Xcal$ whose edges are given by a function $N: \Xcal \rightarrow 2^{\Xcal}$: for each vertex $x \in \Xcal$, $N(x)$ is the set of outgoing neighbors of $x$. We write $N^{-1}(x) := \{y \in \Xcal : x \in N(y)\}$ for the set of in-neighbors of $x$. Given such a function $N$, we consider functions $\dde$ of the form
$$
\dde(S) = \bigcup_{i \in [n]} N(x_i).
$$
On a sample $S \sim D^n$, $\dde$ returns the set of all outgoing neighbors across the examples in $S$, and thus $D(\dde(S))$ measures the mass placed by $D$ on all the neighbors of the sample.
The following theorem shows that for \emph{any} neighboring function $N$, its corresponding $\dde$ function is estimatable.

\begin{theorem} [Unions of Neighborhoods] \label{thm:neighbmass} Let $N: \Xcal \rightarrow 2^{\Xcal}$ be any function and consider the function $\dde(S) = \bigcup_{i \in [n]} N(x_i).$ Then $\dde$ is $O(1/n)$-estimatable via the leave-one-out estimator.
\end{theorem}

\new{Complements require no separate argument. For every mapping $\dde$, the leave-one-out estimate of the complement $\dde^c(S):=\Xcal\setminus\dde(S)$ is $1-\phat^{\mathrm{loo}}_\dde(S)$, and $\gamma_{\dde^c}=\gamma_\dde$ because $A^c\,\Delta\,B^c=A\,\Delta\,B$. Hence $\dde^c$ is estimatable via the leave-one-out estimator at the same rate as $\dde$, and we obtain the following corollary.}

\begin{corollary} \label{cor:neighcomp} Let $N: \Xcal \rightarrow 2^{\Xcal}$ be any function and consider the function $\dde(S) =  \bigcap_{i \in [n]} N(x_i).$ Then $\dde$ is $O(1/n)$-estimatable \new{via the leave-one-out estimator}.
\end{corollary}
\new{Indeed, $\dde(S)^c=\bigcup_{i\in[n]}N'(x_i)$ with $N'(x):=\Xcal\setminus N(x)$, so this follows from Theorem~\ref{thm:neighbmass} and the observation above.}
Note that a special case of Corollary~\ref{cor:neighcomp} is the classical missing mass problem $\dde(S) = S^c.$ Hence, these results provide, to the best of our knowledge, an alternative proof of the fact that missing mass is $O(1/n)$-estimatable.

\paragraph{Estimating Mass of Pierced Sets.} \label{sec:pierce}
In many applications, one is interested in estimating the mass of in-sample examples that satisfy certain properties. For example, in ecology, one might be interested in estimating the mass of species with a \emph{predetermined} anomaly \citep{good1953population, chao1992estimating, bunge2014estimating}. Motivated by these applications, we consider functions $\dde$ of the form
$$\dde(S) = S \cap V$$
where $V \subseteq \Xcal$ is a prespecified ``piercing'' set that is \emph{fixed}, i.e.\ does not depend on the realized samples $S$. Since $S \cap V$ is the union of the neighborhoods $N(x) = \{x\} \cap V$, Theorem~\ref{thm:neighbmass} shows that estimation is then possible at a rate of $O(1/n)$. The set $V$ may itself be chosen as a function of the sample size, say $V = V_n$. The proof is unchanged, but the estimator must then be the \emph{level-$n$} leave-one-out estimator $\frac1n\sum_{i=1}^n \mathbf 1\{x_i \in S^{-i}\cap V_n\}$, which holds $V_n$ fixed across the folds.

\new{The same bound holds for the unseen part $S^c\cap V$ of a fixed region, which is the mass of the \emph{missing} examples with the property, and for the unions $S\cup V$ and $S^c\cup V$.}

\begin{corollary}[\new{Pierced Sets}] \label{thm:strpierc}\label{thm:pierccomp}\label{cor:compstrpierc}
\new{Let $V \subseteq \Xcal$ be any fixed set. Each of the mappings $S\mapsto S\cap V$, $S\mapsto S^c\cap V$, $S\mapsto S\cup V$ and $S\mapsto S^c\cup V$ has $\gamma_\dde(n)\leq 1/n$ for every $n\geq1$, and is therefore $O(1/n)$-estimatable via the leave-one-out estimator.}
\end{corollary}

\begin{proof}
\new{For each of the four mappings, $\dde(S)\,\Delta\,\dde(S^{-n})\subseteq\{x_n\}\setminus T(S^{-n})$, since adding $x_n$ to $S^{-n}$ changes the set of sample points only by adding $x_n$, and only if $x_n\notin T(S^{-n})$. Hence
$$\gamma^D_\dde(n)\leq\Ex\left[D(x_n)\,\mathbf 1\{x_n\notin T(S^{-n})\}\right]=\sum_{x\in\Xcal}D(x)^2\,(1-D(x))^{n-1}\leq\max_{x\in\Xcal}D(x)(1-D(x))^{n-1}\leq\frac1n,$$
and Theorem~\ref{thm:looest} applies.}
\end{proof}

\new{With $V=\emptyset$, the mapping $S\mapsto S^c\cup V$ is the missing mass.}

\paragraph{Estimating Mass of Functions of Heavy Hitters.}
\label{sec:heavy}
One important generalization of missing mass estimation is estimating the mass of heavy hitters. Given a sample $S$ and an integer $k \geq 0$, define $c_k(S)$ to be the set of examples $x \in \Xcal$ that occur exactly $k$ times in $S$ (i.e.\ the $k$-heavy hitters); for $k \geq 1$ we have $c_k(S)\subseteq S$, while $c_0(S) = S^c$.  When $\dde(S) = c_k(S)$, our estimation task reduces exactly to that of estimating the $k$-heavy hitters, which is known to be $O(k/n)$-estimatable \citep{mcallester2000convergence}. Lemma~\ref{lem:khh} shows that $\gamma_\dde(n) \leq \frac{2k+1}{n}$ when $\dde(S) = c_k(S).$ Hence, by Theorem~\ref{thm:looest}, we recover this known result.

\begin{lemma} \label{lem:khh} Fix $k \geq 1$ and let $\dde(S) = c_k(S)$. Then, $\gamma_\dde(n) \leq \frac{2k+1}{n}.$
\end{lemma}

\new{Next, we consider estimating the mass of the neighbors of the $k$-heavy hitters. Unlike Lemma~\ref{lem:khh}, this result has an additional $\log n$ term, and we leave whether the rate $O(k/n)$ holds for every neighborhood function as an open question.}

\begin{theorem} [Unions of Heavy-Hitter Neighborhoods] \label{thm:heavy-hitters} \new{Let $N: \Xcal \rightarrow 2^{\Xcal}$ be any function. For $k \geq 1$, consider the function $\dde(S) = \bigcup_{x \in c_k(S)} N(x).$ Then $\gamma_\dde(n)\leq\frac{3k+2\ln n+2}{n}$ for every $n\geq2$, and $\dde$ is $O\big(\frac{k+\log n}{n}\big)$-estimatable via the leave-one-out estimator.}
\end{theorem}

Finally, we also extend these results to estimating the mass of arbitrary piercings of the $k$-heavy hitters.

\begin{theorem} [Piercings of Heavy Hitters] \label{thm:heavy-hitters-piercing} Let $V \subseteq \Xcal$ be any fixed set. For $k \geq 1$, consider the function $\dde(S) = V \cap c_k(S).$ Then $\dde$ is $O(k/n)$-estimatable via the leave-one-out estimator.
\end{theorem}

\section{Estimating the Error of a Memorizing Classifier via Adaptive Missing Mass Estimation} \label{sec:beyond}

Estimating the true error of a trained classifier is the basic instance of our problem. Most theoretical approaches to it rely on the empirical estimate together with an upper bound on the generalization gap, obtained by one of the methods discussed in Section~\ref{sec:empest}. Yet in practice the empirical error is close to $0$ while the generalization gap is large, often over 20\% \citep{zhang2017understanding}. An estimate obtained in this way cannot be more accurate than the gap it bounds, and so these approaches cannot give a useful estimate of the true error.

In this section we show that by combining the two types of estimators one can estimate the error of a natural class of \emph{exception-memorizing} predictors. These predictors combine a structured base classifier with data-dependent corrections, such as the memorization of individual, atypical training examples. Such classifiers are used both in theory (e.g.,~\citealp{boardpitt90,angluin1997malicious,manoj2023shortest}), most notably in the classical proof that PAC learning implies an Occam (or consistent and compressing) classifier \citep{blumer1987occam,boardpitt90}, and in many practical systems. For example, Batchwise Patching retains an immutable black-box classifier and invokes local patch classifiers selected by an error-region detector for concept drift and transfer learning \citep{kauschke2018batchwise}, and \textsc{ResMem} augments an existing prediction model (e.g., a neural network) by fitting the model's residuals with a nearest-neighbor based regressor \citep{yang2023resmem}. Other correction modules populated by specific errors, edits, or user feedback include feedback-memory graph repair, memory-assisted prompt correction, discrete edit codebooks, per-edit experts, and routed side memories \citep{tandon2022learning,mitchell2022memory,madaan2022memory,hartvigsen2023aging,yao2024scalable,wang2024wise}. It is also known that standard training methods memorize such atypical examples \citep{feldman2020does,feldman2020neural}, although in this case there is typically no explicit separation between the base classifier and the exceptions.

We first describe this class of learners formally, starting with the simplest one, which memorizes every exception in the training set.

\subsection{Exception-Memorizing Learners}

Let inputs come from a space $\Wcal$ with labels given by a fixed labeling function $y : \Wcal \to \Ycal$, and identify the example space with the set of correctly labeled pairs, $\Xcal = \{(w,y(w)) : w \in \Wcal\}$; we suppress the label and write $w$ for the example $(w,y(w))$. A \emph{base learner} $\Acal$ is an algorithm that maps a sample $S$ to a classifier $h_S : \Wcal \to \Ycal$. We write
$$
E(S) := \{(w,y(w)) \in \Xcal : h_S(w) \neq y(w)\}
$$
for the error region of $h_S$. (The base learner may be randomized; for simplicity we suppress the dependence on its random seed and treat it as deterministic.)

The simplest exception-memorizing learner runs $\Acal$ and then memorizes the training set: on an example that occurs in $S$ it predicts the label recorded in $S$, and elsewhere it predicts as $h_S$. Its memorized answers are correct, so its error region is
\begin{equation} \label{eq:memorizing-error}
\dde(S) = E(S) \setminus T(S),
\end{equation}
where $T(S) = \{x_1,\dots,x_n\}$ is the set of sample points, as in Section~\ref{sec:prelim}.

\ifarxiv
In practice, a memorized exception is used to correct the predictions on other, sufficiently similar inputs as well. To model this we use a neighborhood function $N : \Xcal \to 2^{\Xcal}$ as in Section~\ref{sec:neighbors}, which assigns to each example $x$ a set $N(x)$ of neighbors; we assume throughout that $x \in N(x)$. Memorizing an exception $x$ then corrects the base classifier on all of $N(x)$.

\begin{definition}[Neighborhood-correcting learner $\Acal_N$] \label{def:corrected}
Let $\Acal$ be a base learner and $N$ a neighborhood function. For a region $V \subseteq \Xcal$ and a sample $U$, the \emph{correction set}
\begin{equation} \label{eq:correction-rule}
C_V(U) := V \cap \bigcup_{x \in V \cap T(U)} N(x)
\end{equation}
is the part of $V$ that is corrected by memorizing the exceptions in $U$, that is, the points of $U$ that lie in $V$. The learner $\Acal_N$ runs $\Acal$ on $S$ to obtain $h_S$ and then corrects $h_S$ on every neighbor of every exception in $S$. Its error region is
\begin{equation} \label{eq:corrected-error}
\dde(S) := E(S) \setminus C_{E(S)}(S).
\end{equation}
\end{definition}

Here we model the correction as exact: on every neighbor of a memorized exception the corrected classifier is right. Since $C_V(U) \subseteq V$, a correction never introduces an error; it only removes errors of $h_S$. When $N(x) = \{x\}$ we have $C_V(U) = V \cap T(U)$, and $\Acal_N$ is the exception-memorizing learner of \eqref{eq:memorizing-error}, which we denote by $\Acal_{\mathrm{mem}}$.
\fi

Neither of the two standard estimators is adequate for estimating the error of the exception-memorizing classifier. The empirical estimator is identically zero, since $\dde(S)$ excludes every sampled point. The leave-one-out estimator retrains $\Acal$ on each of the $n$ folds, which is expensive, and it is not covered by the guarantee of Theorem~\ref{thm:looest} when $\Acal$ is unstable, for example because its training objective is non-convex; Appendix~\ref{sec:hybrid-separation} gives a simple example in which it fails.

\subsection{Adaptive Missing Mass Estimation} \label{sec:amm}
The true error of the exception-memorizing classifier, $D(\dde(S)) = D(E(S)\setminus T(S))$, is exactly the \emph{missing mass of the error region}. For a region fixed in advance this is the mass of the out-of-sample pierced set we defined in Section~\ref{sec:pierce}, which is $O(1/n)$-estimatable by Corollary~\ref{thm:pierccomp}. In our problem the region $E(S)$ is selected using the same sample whose missing mass is then estimated. This gives rise to a general question of {\em adaptive missing mass estimation} that we formulate below.
\begin{definition}[Adaptive missing mass estimation] \label{def:adaptive-missing-mass}
A \emph{region selection rule} is a map $\Vcal : \Xcal^n \to 2^{\Xcal}$, possibly randomized using an internal seed $r \sim R$, as in Definition~\ref{def:est-general}. For a region $V \subseteq \Xcal$ and a sample $S$, let
\begin{equation} \label{eq:adaptive-missing-mass-value}
M_{V}(S) := D\big(V \setminus T(S)\big)
\end{equation}
be the missing mass of $V$. Adaptive missing mass estimation for a selection rule $\Vcal$ is the problem of estimating $M_{\Vcal(S)}(S)$, the missing mass of the selected region.
\end{definition}

\subsection{The Hybrid Estimator}

The missing mass of the selected region is the difference of two masses,
\begin{equation} \label{eq:target-difference}
M_{\Vcal(S)}(S) = D(\Vcal(S)) - D\big(\Vcal(S)\cap T(S)\big),
\end{equation}
and we estimate the two terms using two different estimators. The mass of the selected region is estimated empirically, by the fraction $\hat D(\Vcal(S))$ of the sample that it contains, which for the error region of a base classifier is its training error. The mass of the seen part is estimated by leave-one-out with the region held fixed: for a fixed region $V$, the leave-one-out estimator of the in-sample piercing $U \mapsto V \cap T(U)$ of Section~\ref{sec:pierce} is
\begin{equation} \label{eq:correction-loo}
\phat^{\mathrm{loo}}_{V \cap T}(S) := \frac1n\sum_{i=1}^n \mathbf 1\big\{x_i \in V \cap T(S^{-i})\big\},
\end{equation}
and the hybrid evaluates it at $V = \Vcal(S)$, the region output on the full sample. Writing $\#_x(S) := \sum_{i=1}^n \mathbf 1\{x_i = x\}$ for the number of occurrences of $x$ in $S$, a point $x_i$ lies in $T(S^{-i})$ exactly when $\#_{x_i}(S)\geq2$, so the difference of the two estimates is a count of singletons.

\begin{definition}[Hybrid estimator] \label{def:hybrid}
For a region selection rule $\Vcal$, the \emph{hybrid estimator} of the missing mass $M_{\Vcal(S)}(S)$ is
\begin{align}
\phat_{\Vcal}^{\mathrm{hyb}}(S)
&:= \frac1n\sum_{i=1}^n\mathbf 1\{x_i\in \Vcal(S)\}
-\frac1n\sum_{i=1}^n\mathbf 1\big\{x_i\in \Vcal(S) \cap T(S^{-i})\big\} \label{eq:hybrid}\\
&= \frac1n\sum_{i=1}^n\mathbf 1\{x_i\in \Vcal(S),\ \#_{x_i}(S)=1\}. \label{eq:hybrid-is-amm}
\end{align}
For the exception-memorizing classifier, $\dde(S) = E(S)\setminus T(S)$, we write $\phat_\dde^{\mathrm{hyb}} := \phat_{E}^{\mathrm{hyb}}$.
\end{definition}

The right-hand side of \eqref{eq:hybrid-is-amm} is the Good--Turing estimator of the missing mass restricted to the selected region. For a region $V$ fixed in advance we write $\widehat M_V(S):=\frac1n\sum_{i=1}^n\mathbf 1\{x_i\in V,\ \#_{x_i}(S)=1\}$ for this \emph{restricted Good--Turing estimator}, so that $\phat_{\Vcal}^{\mathrm{hyb}}(S) = \widehat M_{\Vcal(S)}(S)$. Since $x_i \in V \setminus T(S^{-i})$ exactly when $x_i \in V$ is a singleton, $\widehat M_V$ is the leave-one-out estimator of the out-of-sample piercing $U\mapsto V\setminus T(U)$, and Corollary~\ref{thm:pierccomp} gives $\sup_D\Ex[(\widehat M_V(S)-M_V(S))^2]=O(1/n)$. We note that for the error region of a base classifier, computing the hybrid requires running the base learner once rather than $n$ times.

The two components of the hybrid have errors of different kinds; the following lemma, immediate from \eqref{eq:target-difference} and \eqref{eq:hybrid}, fixes the notation for them.

\begin{lemma}[Error decomposition] \label{lem:hybrid-decomp}
For every region selection rule $\Vcal$ and every sample $S$,
\begin{equation} \label{eq:hybrid-error-identity}
\phat_{\Vcal}^{\mathrm{hyb}}(S)-M_{\Vcal(S)}(S)
=\Big(\hat D(\Vcal(S))-D(\Vcal(S))\Big)
-\Big(\phat^{\mathrm{loo}}_{\Vcal(S)\cap T}(S)-D\big(\Vcal(S)\cap T(S)\big)\Big).
\end{equation}
Consequently, writing
\begin{align}
\varepsilon^{\mathrm{emp}}_{\Vcal}(n) &:= \sup_{D\in\Delta\Xcal}\Ex_{S\sim D^n}\Big[\big(\hat D(\Vcal(S))-D(\Vcal(S))\big)^2\Big], \label{eq:hybrid-empirical-error}\\
\varepsilon^{\mathrm{loo}}_{\Vcal}(n) &:= \sup_{D\in\Delta\Xcal}\Ex_{S\sim D^n}\Big[\big(\phat^{\mathrm{loo}}_{\Vcal(S)\cap T}(S)-D\big(\Vcal(S)\cap T(S)\big)\big)^2\Big], \label{eq:correction-error}\\
\varepsilon^{\mathrm{amm}}_{\Vcal}(n) &:= \sup_{D\in\Delta\Xcal}\Ex_{S\sim D^n}\Big[\big(\phat_{\Vcal}^{\mathrm{hyb}}(S)-M_{\Vcal(S)}(S)\big)^2\Big] \label{eq:amm-error}
\end{align}
for the worst-case mean-squared errors of the empirical estimate of the selected region, of the leave-one-out estimate of its seen part, and of the hybrid, we have
\begin{equation} \label{eq:hybrid-decomp}
\varepsilon^{\mathrm{amm}}_{\Vcal}(n) \leq 2\,\varepsilon^{\mathrm{emp}}_{\Vcal}(n)+2\,\varepsilon^{\mathrm{loo}}_{\Vcal}(n).
\end{equation}
\end{lemma}

The first term is controlled by any of the conditions of Section~\ref{sec:empest}, and the second would be $O(1/n)$ if the region were fixed in advance. The key problem is the adaptivity: the estimate is evaluated at the region $\Vcal(S)$ chosen by the same sample $S$, and this dependence invalidates the standard analysis. One might hope that the generalization of the selection rule is sufficient to control the error: if $\hat D(\Vcal(S))$ is an accurate estimate of $D(\Vcal(S))$, perhaps $\widehat M_{\Vcal(S)}(S)$ is an accurate estimate of $M_{\Vcal(S)}(S)$ as well. The following proposition, proved in Appendix~\ref{app:beyond-proofs}, shows that it is not.
\begin{proposition}[Generalization of the base learner does not suffice for the hybrid estimator] \label{prop:gap-not-enough}
There is a base learner $\Acal$, with a symmetric error-region rule $S\mapsto E(S)$, such that $\varepsilon^{\mathrm{emp}}_{E}(n)=O(1/n)$, while for the exception-memorizing learner built from it $D(\dde(S))=0$ for every sample and
$$
\sup_{D\in\Delta\Xcal}\Ex_{S\sim D^n}\Big[\big(\phat_\dde^{\mathrm{hyb}}(S)-D(\dde(S))\big)^2\Big]
=\varepsilon^{\mathrm{amm}}_{E}(n)\geq c
\qquad\text{for every } n,
$$
for an absolute constant $c>0$.
\end{proposition}

In the construction, $E(S)$ is the set of points that occur exactly once in $S$, provided that the sample passes a consistency check, and it is empty otherwise. The check compares the fraction of the sample made up of such points with the Good--Turing estimate of their mass. Passing this check guarantees that the training error of $h_S$ is an accurate estimate of the population error $D(E(S))$, under every distribution. We show that the check holds with constant probability under the uniform distribution on $n$ points. Yet $E(S)$ contains no unseen point, so the memorizing classifier is perfect, whereas the hybrid counts every singleton of $E(S)$ and reports an error of about $1/e$.

The proposition shows that the two ingredients of the hybrid cannot be analyzed separately: no bound of the form $\varepsilon^{\mathrm{amm}}_{\Vcal}(n)\leq C\,\varepsilon^{\mathrm{emp}}_{\Vcal}(n)+O(1/n)$ holds for all selection rules. We note that in this construction the error of the memorizing classifier is identically zero and is therefore trivially estimatable, so the proposition is a statement about the hybrid estimator and not about estimatability.
We next give two natural conditions on the selection rule under which the error can be controlled. The first bounds the error by the amount of information that the selected region carries about the sample. The second bounds it by the complexity of the class of regions that the rule can output, with no restriction on how the region is chosen from the class.

\begin{theorem}[Adaptive missing mass estimation] \label{thm:adaptive-missing-mass}
Let $\Vcal$ be a region selection rule and let $\Hcal_n$ be the class of all regions that $\Vcal$ can output on samples of size $n$. The constants in the following bounds are absolute.
\begin{enumerate}[label=(\roman*)]
  \item\label{item:amm-info} For every $D\in\Delta\Xcal$, writing $I_D(S;\Vcal)$ for the mutual information (in nats) between the sample $S \sim D^n$ and the selected region,
  $$
  \Ex_{S\sim D^n}\Big[\big(\widehat M_{\Vcal(S)}(S)-M_{\Vcal(S)}(S)\big)^2\Big]
  = O\Big(\frac{I_D(S;\Vcal)+1}{n}\Big).
  $$
  In particular, if $\Vcal$ takes at most $L$ values then $\varepsilon^{\mathrm{amm}}_{\Vcal}(n)=O\big((\log L+1)/n\big)$.
  \item\label{item:amm-vc} If $d_n:=\operatorname{VC}(\Hcal_n)$ satisfies $1\leq d_n\leq n$, then
  $$
  \varepsilon^{\mathrm{amm}}_{\Vcal}(n)
  = O\Big(\frac{d_n\log(en/d_n)}{n}\Big).
  $$
\end{enumerate}
In particular, $\varepsilon^{\mathrm{amm}}_{\Vcal}(n)\to0$ whenever $\sup_{D}I_D(S;\Vcal)=o(n)$ or $d_n=o(n)$.
\end{theorem}

The proof of part~\ref{item:amm-info}, given in Appendix~\ref{app:information-missing-mass}, uses known concentration results for the missing mass \citep{mcallester2003concentration} to show that for every region fixed in advance, the error of the restricted Good--Turing estimator is sub-Gaussian at scale $1/\sqrt n$. It then uses the Donsker--Varadhan variational inequality to decouple the selected region from the sample at a price of $I_D(S;\Vcal)$. This approach is common in information-theoretic generalization bounds mentioned in Section~\ref{sec:related-known}. The information-theoretic argument immediately implies a cardinality-based bound which can also be derived via the union bound from the concentration results of \citet{mcallester2003concentration}. To derive part~\ref{item:amm-vc}, proved in Appendix~\ref{app:vc-missing-mass}, we rely on the cardinality-based bound together with standard covering arguments based on the VC dimension.
\ifarxiv
The two conditions are incomparable. The class of all singletons has VC dimension $1$, yet the rule \new{$\Vcal(S)=\{\min T(S)\}$ on $\Xcal=\mathbb{N}$, which is symmetric,} can reveal arbitrarily many nats about the sample, so \ref{item:amm-vc} is not a consequence of \ref{item:amm-info}; conversely, a rule whose output carries little information about the sample may range over a class of large VC dimension.
\fi

\paragraph{Implications for the error of a memorizing classifier.}
Theorem~\ref{thm:adaptive-missing-mass} can be directly translated into analogous bounds on the error of the hybrid estimator for the exception-memorizing classifier, with the VC dimension of the error-region class $\Ecal_n := \{E(S) : S \in \Xcal^n\}$ replacing that of $\Hcal_n$, and $|\Ecal_n|$ replacing $L$. The condition $\operatorname{VC}(\Ecal_n)=o(n)$ is the same one that makes the error of the base learner itself estimatable by the empirical estimator (Corollary~\ref{cor:empvc}). The theorem shows that it also makes the error of the memorizing learner built on top of it estimatable, via the hybrid, even though for this learner the empirical estimator is identically zero.

Finally, we observe that the VC dimension of the error-region class is bounded for every base learner whose output is constrained to a hypothesis class of bounded VC dimension, for every possible labeling of the data. To state this uniformly over labelings, let the base learner $\Acal$ be defined on all labeled samples in $(\Wcal\times\Ycal)^n$, so that it applies to every labeling $y:\Wcal\to\Ycal$, and write $\Xcal_y:=\{(w,y(w)):w\in\Wcal\}$, \new{$E^y(S):=\{(w,y(w))\in\Xcal_y:h_S(w)\neq y(w)\}$}, $\Ecal_n^y:=\{E^y(S):S\in\Xcal_y^n\}$, and $\dde_y(S):=E^y(S)\setminus T(S)$ for the example space, the error regions of $\Acal$, and the error region of the exception-memorizing learner under the labeling $y$.

\begin{corollary}[Base learners with a hypothesis class of bounded VC dimension] \label{cor:vc-hypothesis-class}
Let $\Ycal=\zo$, let $\Hcal\subseteq\zo^{\Wcal}$ be a class with $\operatorname{VC}(\Hcal)=d$, and let $\Acal$ be a base learner whose output lies in $\Hcal$ on every labeled sample. Then $\operatorname{VC}(\Ecal_n^y)\leq d$ for every labeling $y$ and every $n$. Consequently, if $1\leq d\leq n$, then for every labeling $y$ the hybrid estimator for the error of the exception-memorizing learner built from $\Acal$ satisfies
$$
\sup_{D\in\Delta\Xcal_y}\Ex_{S\sim D^n}\Big[\big(\phat_{\dde_y}^{\mathrm{hyb}}(S)-D(\dde_y(S))\big)^2\Big]
=O\Big(\frac{d\log(en/d)}{n}\Big),
$$
with an absolute constant that does not depend on $y$.
\end{corollary}

The proof (in Appendix~\ref{app:vc-hypothesis-class}) follows directly from the definition of VC dimension. The indicator of the error region of $h$ relative to $y$ is $w\mapsto h(w)\oplus y(w)$, with $\oplus$ addition modulo $2$, and adding $y$ pointwise is a bijection on the labelings of any finite set, so a set is shattered by the error regions of $\Hcal$ if and only if it is shattered by $\Hcal$. Note that the labeling need not belong to $\Hcal$, and the estimator is the same for every labeling, since \eqref{eq:hybrid-is-amm} is computed from the labeled sample and $h_S$ alone. The corollary extends to an arbitrary label set, with $\operatorname{VC}(\Hcal)$ replaced by the graph dimension of $\Hcal$; see Corollary~\ref{cor:vc-hypothesis-class-multi} in Appendix~\ref{app:vc-hypothesis-class}.

\ifarxiv
\new{For the neighborhood-correcting learner $\Acal_N$, bounding the error of the hybrid estimator beyond the memorization case is an open problem, discussed in Section~\ref{sec:discussion}.}
\fi

\section{The Limits of Estimatability} \label{sec:limits}
Our framework leads to a natural question: when is a data-dependent function $\dde$ estimatable by \emph{some} procedure? We first observe that standard lower bounds on learnability of function classes or distributions do not directly imply the existence of non-estimatable mappings, nor do classical estimation lower bounds (Appendix~\ref{app:classnequnclass}). In these lower bounds, one usually proves that the error of the learning process is high with high probability. Such a lower bound typically defines an event (or set corresponding to the loss) whose probability is high and, importantly, known. We demonstrate that there exist ways to define functions $\dde$ in the transition between the learnable regime and the hard regime so that the defined event is not estimatable.

We give three constructions of non-estimatable events, of decreasing domain size, which we include for completeness: the first and simplest uses a domain of size $2^{2^{n}+1}$, the second $2^{n+2}$, and the third only $O(n^2)$. The first construction is a simple instance of post-selection which shows that even a simple event that selects the coordinate with the largest empirical mean cannot be estimated when the number of coordinates is exponentially large in $n$. The second construction corresponds to learning parity functions and is close to the constructions of \citet{gastpar2024fantastic}, and the third is a random code, as is the construction underlying Theorem~2.1 of \citet{gastpar2024which}. Since both are stated directly in terms of events, their proofs are self-contained and, we believe, somewhat simpler. The mechanism in all three constructions is the same. With constant probability the sample is consistent with a small number of candidate distributions that are equally likely given the sample, and the selected event has different masses under them, so that no estimator can do better than the posterior variance of the target.

\begin{theorem}
\label{thm:max-selection-lb}
For every $n\geq 1$, let $d=2^n+1$ and $\Xcal_n=\{0,1\}^d$. For $S\in\Xcal_n^n$, let $i(S)$ be the smallest coordinate $j\in[d]$ maximizing $\widehat{\mu}_j(S)=\frac{1}{n}\sum_{t=1}^n x_{t,j}$, and define $$\dde_{\max}(S)=\{x\in\Xcal_n:x_{i(S)}=1\}.$$ Then no estimator achieves mean-squared error below $1/64$ for $\dde_{\max}$.
\end{theorem}

\begin{proof}%
For each $j\in[d]$, let $D_j$ be the product distribution on $\Xcal_n$ in which coordinate $j$ is deterministically $1$ and every coordinate $k\neq j$ is an independent $\mathrm{Bern}(1/2)$ random variable.  For these product laws, $D_j(\dde_{\max}(S))$ is exactly the mean of the coordinate selected by the maximum empirical mean rule.

Fix any estimator $\phat:\Xcal_n^n\to[0,1]$.  It suffices to lower-bound the Bayes risk when $J$ is uniform on $[d]$ and, conditional on $J$, $S\sim D_J^n$.  Given the sample, let
\[
A(S)=\{k\in[d]:x_{t,k}=1\ \text{for all }t\in[n]\}
\]
be the set of all-one coordinates.  Since $\widehat\mu_j(S)=1$ exactly when $j\in A(S)$, and $A(S)\neq\emptyset$ because $J\in A(S)$, the maximum empirical mean equals $1$ and its argmax set is $A(S)$; hence $i(S)=\min A(S)$.  Moreover, each $k\neq J$ lies in $A(S)$ independently with probability $2^{-n}$.  Let $E$ be the event $|A(S)|=2$.  Then
\[
\Pr[E]=(d-1)2^{-n}(1-2^{-n})^{d-2}
=(1-2^{-n})^{2^n-1}
\geq \frac14,
\]
where the last inequality holds because $(1-2^{-n})^{2^n-1}$ is decreasing in $n$ with limit $e^{-1}>1/4$.

Now fix a realization $S$ for which $E$ holds and write $A(S)=\{a,b\}$ with $a<b$.  By symmetry,
\[
\Pr[J=a\mid S]
=
\Pr[J=b\mid S]
=\frac12.
\]
On this realization, $i(S)=a$.  Therefore
\[
D_J(\dde_{\max}(S))
=
\begin{cases}
1, & \mbox{if }J=a,\\
1/2, & \mbox{if }J=b.
\end{cases}
\]
Conditioned on $S$ and $E$, the target is thus uniform on $\{1,1/2\}$, so every estimate has conditional squared loss at least its variance, namely $1/16$.  Averaging,
\[
\Ex_{J,S}
\left[
\left(\phat(S)-D_J(\dde_{\max}(S))\right)^2
\right]
\geq
\Pr[E]\cdot\frac1{16}
\geq
\frac1{64}.
\]
Since the supremum over $D\in\Delta\Xcal_n$ is at least this average over $D_1,\ldots,D_d$, the claim follows.
\end{proof}

A limitation of the lower bound in Theorem~\ref{thm:max-selection-lb} is that it requires an input space of size $2^{2^{\Omega(n)}}$.  Next, we give another example of a hard-to-estimate event that directly corresponds to learning parity functions in dimension $n+2$.  This example is still relatively easy to analyze and the input space size is exponentially smaller: $2^{n+2}$.
\begin{theorem}
\label{thm:lb-restated}
For every $n\in\mathbb{N}$, let $\Xcal=\mathbb{F}_2^{n+2}$.  For a sample $S\in\Xcal^n$, let $$V_S=\{w\in\mathbb{F}_2^{n+2}:\langle x_i,w\rangle=0\ \text{for all }i\in[n]\},$$ let $z_S$ be the lexicographically first non-zero vector in $V_S$ if $V_S\neq\{0\}$, and $z_S=0$ otherwise, and define $\dde(S)=\{x\in\Xcal:\langle x,z_S\rangle=0\}$.  Then no estimator achieves mean-squared error below $1/36$ for $\dde$.
\end{theorem}

\begin{proof}[Proof of Theorem~\ref{thm:lb-restated}]
Fix $n \in \mathbb{N}$ and let $\Xcal = \mathbb{F}_2^{n+2}$.
For any vector $w \in \mathbb{F}_2^{n+2}$, define $w^{\perp} := \{u \in \mathbb{F}_2^{n+2}\mid \langle u, w \rangle = 0\}$, i.e., the subspace orthogonal to $w$, which has dimension $n+1$ (whenever $w \neq 0$).
The function $\dde$ is defined in the theorem statement.
Note that both $\Xcal$ and $\dde$ depend on $n$. This still yields a single non-estimatable mapping on a fixed countable domain.
Concretely, one may take $\Xcal=\zo^\star$, the (countable) set of all finite binary strings, and apply the construction to the projection of each sample point onto its first $n+2$ coordinates, padding with zeros as needed.

Next, consider the following distribution over input distributions (which may depend on $n$). Choose a non-zero vector $z \in \mathbb{F}_2^{n+2}$ uniformly at random, and let $D_z$ be the uniform distribution on $z^{\perp}.$ Now, let $\phat: \Xcal^n \to [0, 1]$ be any purported estimator. We will show that
\[
\Ex_{z, S}[(\phat(S) - D_z(\dde(S)))^2] \geq c := \tfrac{1}{36},
\]
which implies that $\phat$ is not a constant-error estimator for $\dde$.

To prove this, first note that under $D_z$, each $x_i$ is uniform on the $(n+1)$-dimensional space $z^{\perp}$, so the probability that $S$ is linearly independent is
\[
p := \left(1-\frac{1}{2^{n+1}}\right)\left(1-\frac{2}{2^{n+1}}\right)\cdots \left(1-\frac{2^{n-1}}{2^{n+1}}\right)
\;\geq\; 1 - \frac{1}{2^{n+1}} - \cdots - \frac{2^{n-1}}{2^{n+1}}
= 1 - \frac{2^n-1}{2^{n+1}} \;\geq\; \frac{1}{2}.
\]
Next, given a linearly independent sample set $S$, there are exactly 3 non-zero vectors orthogonal to all elements of $S$, one of which is $z_S$. If the other two non-zero vectors orthogonal to $S$ are $u$ and $v$, then by symmetry,
$\Pr[z = z_S\mid S] = \Pr[z = u\mid S] = \Pr[z = v\mid S] = \frac{1}{3}$. Furthermore, if $z = z_S$, then $\dde(S)$ contains all of $z_S^{\perp}$, so $D_z(\dde(S)) = 1$. If $z \neq z_S$ (i.e., $z\in\{u,v\}$), then $z^{\perp} \cap z_S^{\perp}$ has dimension $n$, so within $z^{\perp}$ we have $D_z(\dde(S)) = \frac{2^n}{2^{n+1}}=\frac{1}{2}$.

Thus conditioned on the linear independence of $S$, the variance of the target value is easily computed to be $1/18$. Since independence holds with probability at least $1/2$, we conclude that $\Ex_{z, S}[(\phat(S) - D_z(\dde(S)))^2] \geq \frac{1}{36}$.
\end{proof}

\subsection{Hardness on a Domain of Quadratic Size}
\label{sec:quadratic}

Both constructions above require a domain that is exponential in $n$.  We now show that non-estimatability already arises on a domain whose size is only \emph{quadratic} in $n$.

\begin{theorem}
\label{claim:quadratic-domain-lb}
\new{For every $n\geq2$ there is a domain $\Xcal_n$ with $|\Xcal_n|=2n^2$} and a function $\dde_n:\Xcal_n^n\to \zo^{\Xcal_n}$ such that
\[
\inf_{\phat:\Xcal_n^n\to[0,1]}
\sup_{D\in\Delta(\Xcal_n)}
\Ex_{S\sim D^n}
\left[
\left(\phat(S)-D(\dde_n(S))\right)^2
\right]
\ge \new{\frac{1}{576}}.
\]
\end{theorem}

\paragraph{Proof outline.}
The full proof is in Appendix~\ref{app:lower-bound-proofs}; we sketch the idea.  \new{Fix $w=n^2$, $m=2w$ and $\Xcal_n=[m]$. We select a random ``code'' of $M$ sets $V_1,\ldots,V_M$, each an independent uniformly random $w$-subset of $[m]$, with $M=\lfloor 1/p_n\rfloor+1$, where $p_n=\Pr[T\subseteq V_j]$ for a fixed $n$-element set $T$ (so $p_n\le 2^{-n}$ and $M\approx 2^n$). Let $D_i$ be uniform on $V_i$, and let $\dde_{\mathcal C}(S)$ output the first set $V_i$ that contains the sampled set $T(S)=\{x_1,\dots,x_n\}$. Consider the Bayesian experiment with $I$ uniform on $[M]$ and $S\sim D_I^n$. With constant probability the sample has no repetitions, and then the number of \emph{other} $V_j$'s containing $T(S)$ is $\mathrm{Binomial}(M-1,p_n)$, which takes the value $1$ with constant probability. In that case there are exactly two candidate supports $V_a$ and $V_b$, the posterior distribution of $I$ is uniform on $\{a,b\}$, and the estimand is $1$ for one of them and $\rho=|V_a\cap V_b|/w$ for the other. The other candidate is a uniformly random $w$-set containing $T(S)$, so $\Ex[1-\rho]=(n-1)/(2n-1)\geq1/3$, and by Jensen's inequality the posterior variance $(1-\rho)^2/4$ is at least $1/36$ on average. Averaging over the code and applying the probabilistic method fixes a single code.}

We remark that in this case the image class $\Hcal_n=\{\dde_{\mathcal C}(S):S\in\Xcal_n^n\}$ has size $M+1\le 2^{n+1}$, so its VC dimension is at most $n+1$.

\section{Discussion and Open Problems}
\label{sec:discussion}

In this work we studied data-dependent probability estimation via a unified framework that incorporates a variety of existing problems and methods, and used it to estimate the error of exception-memorizing predictors via adaptive missing mass estimation. We also described several limitations on estimatability. Our results suggest that estimatability can arise from fundamentally different structural properties. Empirical estimators exploit generalization control, for example through low complexity of the induced image class, algorithmic stability, differential privacy, or information-theoretic bounds, while leave-one-out estimators rely on stability under small perturbations of the sample. Understanding how these principles interact, and whether they admit a unified characterization, remains an intriguing direction for future work. We list a few directions below.

\paragraph{Characterizing estimatability.}
The most fundamental open problem is to characterize which functions $
\dde:\Xcal^\star\to \zo^\Xcal$
are estimatable.
 Our results provide several sufficient conditions and impossibility results, but a complete characterization remains open. In particular, it is unclear whether there exists a single structural notion that captures estimatability, or whether fundamentally different mechanisms are unavoidable.

\paragraph{Domain size of hard instances.}
Theorem~\ref{claim:quadratic-domain-lb} gives a non-estimatable mapping on a domain of size $O(n^2)$. On the other hand, if $|\Xcal_n|=o(n)$, then for \emph{every} $\dde$ the image class $\Hcal_n=\{\dde(S):S\in\Xcal_n^n\}$ has VC dimension at most $|\Xcal_n|=o(n)$, so the empirical estimator is already consistent by Corollary~\ref{cor:empvc}. Closing the gap between this linear lower bound and the quadratic construction is a natural and interesting open problem.

\paragraph{Neighborhood-correcting learners.}
Our results for the exception-memorizing learner concern a classifier that overrides the base classifier only at the training points themselves. In practice a memorized exception is also used to correct the predictions on nearby inputs, and the natural model is a learner that corrects $h_S$ on the neighborhood $N(x)$ of every exception $x$ in the sample, for a neighborhood function $N$ as in Section~\ref{sec:neighbors}\ifarxiv{} (Definition~\ref{def:corrected})\fi. Its error region is $E(S)$ with the part covered by the neighborhoods of the sample points in $E(S)$ removed, and the hybrid estimator extends naturally: the mass of $E(S)$ is estimated by the training error, and the corrected mass by leave-one-out with the region held fixed. For a region fixed in advance the corrected part is a union of neighborhoods, so the second estimate is $O(1/n)$-accurate by Theorem~\ref{thm:neighbmass}. What is missing is the adaptive step. The proof of Theorem~\ref{thm:adaptive-missing-mass} rests on exponential concentration of the error of the restricted Good--Turing estimator for a fixed region, which comes from the concentration of the missing mass \citep{mcallester2003concentration} together with a bounded-differences argument for the count of singletons. Neither ingredient is available once neighborhoods are involved: in general a single sample point can cover or uncover many others, so the leave-one-out estimate of the corrected mass does not have bounded differences, and the indicators of the uncovered points need not be negatively associated, so the argument for the missing mass does not apply. For the uncovered mass itself, concentration is known only under additional conditions in metric spaces \citep{maurer2022concentration}. Establishing such concentration, or bounding the error of the hybrid for neighborhood-correcting learners by other means, in terms of the complexity of the error-region class or the information it carries about the sample, is an open problem.

\ifarxiv
\paragraph{Optimal rates and complexity measures.}
Our analysis primarily focuses on establishing consistency under broad conditions. A natural next step is to determine optimal rates of convergence for different families of functions $\dde$.
\fi

\paragraph{Beyond the current estimators.}
The examples in this work suggest that empirical and leave-one-out estimators are complementary rather than competing approaches and can be combined to obtain new estimators. This raises the possibility of developing new estimators that adaptively combine multiple structural properties of $\dde$. Understanding the space of such estimators, and identifying natural problems that require genuinely new techniques, is an interesting direction for future work.

\subsubsection*{AI Use Disclosure}
The authors formulated the framework, initiated the research on all questions and wrote a preliminary draft of this work. All parts of this version were created via an interactive process in which AI (Claude Opus/Fable;ChatGPT Sol/Astra) generated drafts, details of proofs, and improved authors' proofs based on author-provided structure and instructions. The resulting text was also extensively edited and proofread by the authors. The authors take full responsibility for the content of this paper.

\appendix

\section{Results for the Empirical Estimator} \label{app:defvc}
\begin{definition}[VC dimension \citep{vapnik1971uniform}] \label{def:vc}
Let $\Hcal \subseteq \zo^{\Xcal}$.  A set $\{x_1, \ldots, x_{d}\}\subseteq\Xcal$ is shattered by $\Hcal$ if, for every $y_1, \ldots, y_{d} \in \zo$, there exists $h\in \Hcal$ such that $h(x_i) = y_i$ for all $i \in [d]$. The VC dimension of $\Hcal$, denoted $\operatorname{VC}(\Hcal)$, is the largest integer $d \geq 0$ for which there exists a set $\{x_1, \ldots, x_{d}\}\subseteq\Xcal$ that is shattered by $\Hcal$.
\end{definition}

This section proves Corollary~\ref{cor:empvc} and Theorem~\ref{thm:smallcard} of Section~\ref{sec:empest} and shows that the two complexity measures they involve are incomparable.

\new{For a class $\Fcal$ of subsets of $\Xcal$ we write
$$
\|\hat D-D\|_{\Fcal}:=\sup_{F\in\Fcal}\big|\hat D(F)-D(F)\big|
$$
for the uniform deviation of the empirical measure over $\Fcal$. Both Corollary~\ref{cor:empvc} and the VC bound for adaptive missing mass (Appendix~\ref{app:vc-missing-mass}) rest on the following bound.}

\begin{lemma}[Uniform deviation] \label{lem:amm-uniform}
\new{If $\operatorname{VC}(\Fcal)=v<\infty$ then $\sup_{D\in\Delta\Xcal}\Ex_{S\sim D^n}\big[\|\hat D-D\|_{\Fcal}^2\big]=O\big((v+1)/n\big)$.}
\end{lemma}

\begin{proof}
\new{The classical chaining bound on the Rademacher complexity of a VC class gives $\Ex\|\hat D-D\|_{\Fcal}=O\big(\sqrt{(v+1)/n}\big)$ \citep{talagrand1994sharper,boucheron2013concentration,vershynin2018highdimensional}, and $\|\hat D-D\|_{\Fcal}$ changes by at most $1/n$ when one sample point is replaced, so the bounded-differences form of the Efron--Stein inequality \citep{boucheron2013concentration} gives $\operatorname{Var}(\|\hat D-D\|_{\Fcal}) \le 1/(4n)$. Hence $\Ex\|\hat D-D\|_{\Fcal}^2=(\Ex\|\hat D-D\|_{\Fcal})^2+\operatorname{Var}(\|\hat D-D\|_{\Fcal})=O\big((v+1)/n\big)$, uniformly in $D$.}
\end{proof}

\begin{proof}[Proof of Corollary~\ref{cor:empvc}]
\new{Since $\dde(S)\in\Hcal_n$, the estimation error satisfies $|\phat_\dde^{\text{emp}}(S)-D(\dde(S))|=|\hat D(\dde(S))-D(\dde(S))|\leq\|\hat D-D\|_{\Hcal_n}$, so Lemma~\ref{lem:amm-uniform} with $\Fcal=\Hcal_n$ gives the claim.}
\end{proof}

We record the cardinality-based complexity measure of \eqref{eq:small} for reference:
\begin{equation*}
\tau_\dde(n) := \sup_{D \in \Delta \Xcal} \mathop{\Ex}_{S \sim D^n} |\dde(S)|.
\end{equation*}

\begin{proof}[\new{Proof of Theorem~\ref{thm:smallcard-mse}}]
\new{Fix $D\in\Delta\Xcal$ and write $A=\dde(S)$, $p_x=D(x)$, $\hat p_x=\hat D(x)$ and $Y:=\sum_{x\in\Xcal}(\hat p_x-p_x)^2$. Since $\Ex|A|\leq\tau_\dde(n)<\infty$, the set $A$ is finite almost surely, and Cauchy--Schwarz over $A$ gives
$$\left(\phat_\dde^{\mathrm{emp}}(S)-D(A)\right)^2=\Bigl(\sum_{x\in A}(\hat p_x-p_x)\Bigr)^2\leq |A|\,Y.$$
If $|A|\leq\tau_0$ for every sample, taking expectations and using $\Ex Y=\frac1n\sum_x p_x(1-p_x)\leq\frac1n$ gives the second claim. In general, since the error is at most $1$ in absolute value, for every $y>0$
$$\Ex\left[\left(\phat_\dde^{\mathrm{emp}}(S)-D(A)\right)^2\right]\leq \Ex\left[|A|\,Y\,\mathbf 1\{Y\leq y\}\right]+\Pr[Y>y]\leq y\,\tau_\dde(n)+\Pr\left[\sqrt Y>\sqrt y\right].$$
The function $f(S):=\sqrt{Y}=\|\hat p-p\|_2$ satisfies $\Ex f\leq\sqrt{\Ex Y}\leq1/\sqrt n$, and replacing one sample point changes $\hat p$ by a vector of $\ell_2$ norm at most $\sqrt2/n$, so McDiarmid's inequality gives $\Pr[f>1/\sqrt n+t]\leq e^{-nt^2}$ for every $t>0$. We take $t=\sqrt{\ln(n/\tau_\dde(n))/n}$ and $\sqrt y=1/\sqrt n+t$, so that $y\leq 2/n+2t^2$ and $\Pr[\sqrt Y>\sqrt y]\leq\tau_\dde(n)/n$. This gives
$$\Ex\left[\left(\phat_\dde^{\mathrm{emp}}(S)-D(A)\right)^2\right]\leq\frac{\tau_\dde(n)}{n}\left(2\ln\frac{n}{\tau_\dde(n)}+2\right)+\frac{\tau_\dde(n)}{n},$$
and taking the supremum over $D$ proves the first claim.}
\end{proof}

We end this section by showing that the upper bounds of Corollary~\ref{cor:empvc} and Theorem~\ref{thm:smallcard} are incomparable: writing $\Hcal_{\dde,n} := \{\dde(S) : S \in \Xcal^n\}$ for the image class of a mapping $\dde$, there exist $\dde_1, \dde_2$ such that (1) $\tau_{\dde_1}(n) = \Omega(n)$ but $\operatorname{VC}(\Hcal_{\dde_1, n}) = 0$, and (2) $\tau_{\dde_2}(n) = o(n)$ but $\operatorname{VC}(\Hcal_{\dde_2, n}) = \Omega(n)$.  For (1), take $\Xcal = \mathbb{N}$ and consider $\dde_1(S) = \{1, \dots, n\}$, so that $\tau_{\dde_1}(n) = n$ while $\Hcal_{\dde_1,n}$ is a singleton. For (2), take $\Xcal = \mathbb{N} \times \Pcal_{\text{fin}}(\mathbb{N})$, where $\Pcal_{\text{fin}}(\mathbb{N})$ denotes the collection of finite subsets of $\mathbb{N}$, and consider

$$\dde_2(S) = \begin{cases}
			\{(i, \emptyset): i \in A\}, & \text{if for some $A \subseteq [n]$, the sample is $\{(1, A), \dots, (n, A)\}$}\\
            \emptyset, & \text{otherwise}
		 \end{cases}.$$

Note that $\operatorname{VC}(\Hcal_{\dde_2, n}) = n$ since we can always consider samples $\{(1, A), \dots, (n, A)\}$ for all $A \subseteq [n].$ On the other hand, $\dde_2(S)$ is nonempty only when the $n$ points of $S$ are distinct and exhaust $\{(1,A),\dots,(n,A)\}$ for a single $A$.  Writing $m_A := \sum_{i \in [n]} D((i,A))$, the probability of that event for a given $A$ is $n!\prod_{i \in [n]} D((i,A)) \leq \frac{n!}{n^n} m_A^n$ by the AM--GM inequality; the events for distinct $A$ are disjoint and $\sum_A m_A^n \leq (\sum_A m_A)^n \leq 1$.  Since $|\dde_2(S)| \leq n$ always, $\tau_{\dde_2}(n) \leq \frac{n \cdot n!}{n^n} = \Theta\big(n^{3/2}e^{-n}\big) = o(n)$ by Stirling's approximation.

\section{Missing Proofs for Section~\ref{sec:looest}} \label{app:missproof}
\begin{proof}[Proof of Theorem~\ref{thm:looest}]
\new{At a fixed $D$, this is the bound of \citet{devroye1979distribution} on the deleted estimate in the form of \citet[Lemma~9]{bousquet2002stability}, applied to the loss $\ell(\dde(S),x)=\dde(S)(x)$, whose hypothesis stability \citep[Definition~3]{bousquet2002stability} is $\gamma^D_\dde(n)$ (see the discussion after \eqref{eq:stab}); taking the supremum over $D$ gives the uniform form.}

\end{proof}

\begin{proof}[Proof of Theorem~\ref{thm:neighbmass}] By Theorem~\ref{thm:looest}, it suffices to bound $\gamma_\dde(n).$ Fix $D$ and write $q_x := D(N^{-1}(x))$ for the mass of the in-neighbors of $x$. Deleting $x_n$ from $S$ removes from $\dde(S)$ exactly the points of $N(x_n)$ that lie in no other neighborhood, so for $x \sim D$ independent of $S$,
$$
\Ex_{S \sim D^n}\left[D(\dde(S) \Delta \dde(S^{-n}))\right]
= \Pr\left[x \in N(x_n),\ x \notin N(x_i) \text{ for all } i < n\right]
= \Ex_{x \sim D}\left[q_x(1-q_x)^{n-1}\right] \leq \frac{1}{n},
$$
since $x \in N(x_i)$ if and only if $x_i \in N^{-1}(x)$, these $n$ events are independent with probability $q_x$ given $x$, and $p(1-p)^{n-1} \leq 1/n$ for all $p \in [0,1]$. Thus, $\gamma_\dde(n) \leq \frac{1}{n}$, and hence $\dde$ is estimatable with MSE that decays at a rate of $O(1/n)$.
\end{proof}

\new{Corollary~\ref{thm:strpierc} is proved in the main text. Lemma~\ref{lem:khh} and Theorem~\ref{thm:heavy-hitters-piercing} rest on one identity}: for $p \in [0,1]$ and $0 \leq j \leq n-1$,
\begin{equation} \label{eq:occupancy}
p \binom{n-1}{j} p^{j}(1-p)^{n-1-j} = \frac{j+1}{n}\binom{n}{j+1} p^{j+1}(1-p)^{n-1-j} \leq \frac{j+1}{n},
\end{equation}
which uses $\binom{n-1}{j} = \frac{j+1}{n}\binom{n}{j+1}$ and $\binom{n}{j+1} p^{j+1}(1-p)^{n-(j+1)} = \Pr[\operatorname{Bin}(n,p) = j+1] \leq 1$. In words, the probability that a point of mass $p$ is drawn and occurs exactly $j$ times among $n-1$ further draws is at most $(j+1)/n$.

\begin{proof}[\new{Proof of Lemma~\ref{lem:khh} and Theorem~\ref{thm:heavy-hitters-piercing}}] \new{Both mappings have the form $\dde(S)=\bigcup_{x\in c_k(S)}N(x)$ for a neighborhood function $N$, and we first bound $\gamma_\dde(n)$ for an arbitrary $N$.} Fix $k \geq 1$ and a distribution $D \in \Delta \Xcal$. Adding $x_n$ to $S^{-n}$ increases the multiplicity of $x_n$ by one and leaves every other point unchanged. Hence $x_n$ \emph{enters} $c_k$ exactly when its multiplicity in $S^{-n}$ is $k-1$ and \emph{leaves} $c_k$ exactly when its multiplicity is $k$, and no other point changes status; in either case the union of neighborhoods changes only within $N(x_n)$. Therefore, for any $S \in \Xcal^n$,
$$
D(\dde(S) \Delta \dde(S^{-n})) \leq \mathbf{1}\{x_n \in c_{k-1}(S^{-n}) \cup c_k(S^{-n})\} D(N(x_n)).
$$
Conditioning on $x_n = x$, the multiplicity of $x$ in $S^{-n}$ is distributed as $\operatorname{Bin}(n-1, D(x))$, so by \eqref{eq:occupancy} with $j = k-1$ and $j = k$,
$$
\Ex_{S \sim D^n}\left[D(\dde(S) \Delta \dde(S^{-n}))\right]
\leq \sum_{x \in \Xcal} D(N(x))\, D(x) \Pr\left[\operatorname{Bin}(n-1, D(x)) \in \{k-1, k\}\right]
\leq \frac{2k+1}{n}\sum_{x \in \Xcal} D(N(x)).
$$
For $\dde(S) = c_k(S)$ take $N(x) = \{x\}$: then $\sum_x D(N(x)) = 1$, and the displayed bound reads $\gamma_\dde(n) \leq \frac{2k+1}{n}$, which is Lemma~\ref{lem:khh}. For $\dde(S) = V \cap c_k(S)$ take $N(x) = \{x\} \cap V$: then $\sum_x D(N(x)) = D(V) \leq 1$, so again $\gamma_\dde(n) \leq \frac{2k+1}{n}$, and Theorem~\ref{thm:looest} gives estimation error $O(\frac{k}{n})$.
\end{proof}

\begin{proof}[\new{Proof of Theorem~\ref{thm:heavy-hitters}}]
\new{Fix $n\geq2$, $D\in\Delta\Xcal$ and a point $z\in\Xcal$, let $U:=N^{-1}(z)$, and write $p_u:=D(u)$. Then $z\in\dde(S)$ if and only if $c_k(S)\cap U\neq\emptyset$. Passing from $S^{-n}$ to $S$ raises the multiplicity of $x_n$ by one and changes no other multiplicity, so $c_k(S)$ and $c_k(S^{-n})$ differ at most in $x_n$. Hence $z\in\dde(S)\setminus\dde(S^{-n})$ exactly when $c_k(S)\cap U=\{x_n\}$, and $z\in\dde(S^{-n})\setminus\dde(S)$ exactly when $c_k(S^{-n})\cap U=\{x_n\}$.

By symmetry of $c_k$, $\Pr[c_k(S)\cap U=\{x_n\}]=\frac1n\Ex\sum_{i=1}^n\mathbf 1\{c_k(S)\cap U=\{x_i\}\}$. When $c_k(S)\cap U$ is a single point, exactly $k$ indices $i$ have $x_i$ equal to it, and otherwise the sum is $0$. So this probability is $\frac kn\Pr[|c_k(S)\cap U|=1]\leq\frac kn$.

For the second event, $x_n$ is independent of $S^{-n}$, so with $E_u:=\{c_k(S^{-n})\cap U=\{u\}\}$ its probability is $\sum_{u\in U}p_u\Pr[E_u]$. The events $E_u$ are disjoint, so $\sum_u\Pr[E_u]\leq1$, and $E_u$ requires the multiplicity of $u$ in $S^{-n}$, which has distribution $\operatorname{Bin}(n-1,p_u)$, to equal $k$. Let $T:=(2k+2\ln n)/(n-1)$. The terms with $p_u\leq T$ contribute at most $T$. For $p_u>T$ the mean $\mu:=(n-1)p_u$ exceeds $\mu_0:=2k+2\ln n>k$, and the Chernoff bound $\Pr[\operatorname{Bin}(n-1,p_u)\leq k]\leq\inf_{s>0}e^{sk}\,\Ex e^{-s\operatorname{Bin}(n-1,p_u)}\leq e^{-\mu}(e\mu/k)^k$ (take $e^{-s}=k/\mu$) has a right-hand side that is decreasing in $\mu>k$. At $\mu=\mu_0$, writing $a:=\ln n/k$, it equals $\exp\bigl(k(\ln(2+2a)-1-2a)\bigr)\leq e^{-ak}=1/n$, since $\ln(2+2a)\leq\ln2+a<1+a$. So these terms contribute at most $\frac1n\sum_u p_u\leq\frac1n$.

Therefore $\Pr[z\in\dde(S)\,\Delta\,\dde(S^{-n})]\leq\frac kn+T+\frac1n$ for every $z$, and averaging over $z\sim D$ gives $\gamma^D_\dde(n)\leq\frac{k+1}{n}+\frac{2k+2\ln n}{n-1}$. If $2k+2\ln n\leq n-1$, this is at most $(3k+2\ln n+2)/n$; otherwise $(3k+2\ln n+2)/n>1\geq\gamma^D_\dde(n)$. Taking the supremum over $D$ and applying Theorem~\ref{thm:looest} completes the proof.}
\end{proof}

\section{Proofs for Section~\ref{sec:beyond}}
\label{app:beyond-proofs}

\subsection{Proof of Proposition~\ref{prop:gap-not-enough}}

Recall from Section~\ref{sec:heavy} that $c_k(S)$ is the set of points that occur exactly $k$ times in $S$, and from Remark~\ref{rem:good-turing} that $F_k(S):=|c_k(S)|$. For a sample of size $n$ let
$$
B_n:=\big\{S:\ |F_1(S)-2F_2(S)|\leq6\sqrt n\big\}
$$
and define the region selection rule
$$
E(S):=\begin{cases}c_1(S),& S\in B_n,\\ \emptyset,&\text{otherwise.}\end{cases}
$$
The rule depends on $S$ only through its multiset of points, so it is symmetric, and it is the error region of a base learner: given the labeling $y$, let $h_S$ predict $y(x)$ for $x\notin E(S)$ and a label different from $y(x)$ for $x\in E(S)$. Throughout, $\dde(S)=E(S)\setminus T(S)$ is the error region of the exception-memorizing learner built from this base learner.

\emph{The base learner generalizes.} Every point of $c_1(S)$ occurs once in $S$, so $\hat D(c_1(S))=F_1(S)/n$. By Remark~\ref{rem:good-turing}, the leave-one-out estimator of the mapping $S\mapsto c_1(S)$ is $2F_2(S)/n$, and Lemma~\ref{lem:khh} together with Theorem~\ref{thm:looest} gives
$$
\sup_{D\in\Delta\Xcal}\Ex_{S\sim D^n}\left[\left(\frac{2F_2(S)}{n}-D(c_1(S))\right)^2\right]\leq\frac1{2n}+\frac9n\leq\frac{10}n .
$$
On $B_n$ we have $\hat D(E(S))=F_1(S)/n$ and $D(E(S))=D(c_1(S))$, so
$$
\big|\hat D(E(S))-D(E(S))\big|
\leq\frac{|F_1(S)-2F_2(S)|}{n}+\left|\frac{2F_2(S)}{n}-D(c_1(S))\right|
\leq\frac6{\sqrt n}+\left|\frac{2F_2(S)}{n}-D(c_1(S))\right|,
$$
while outside $B_n$ both $\hat D(E(S))$ and $D(E(S))$ are 0. Squaring with $(u+v)^2\leq2u^2+2v^2$ and taking expectations,
$$
\varepsilon^{\mathrm{emp}}_{E}(n)=\sup_{D\in\Delta\Xcal}\Ex_{S\sim D^n}\left[\left(\hat D(E(S))-D(E(S))\right)^2\right]\leq\frac{72}n+\frac{20}n=\frac{92}n .
$$

\emph{The hybrid fails.} Since $E(S)\subseteq c_1(S)\subseteq T(S)$, we have $\dde(S)=\emptyset$ and $D(\dde(S))=M_{E(S)}(S)=0$ for every sample. On the other hand every point of $E(S)$ is a singleton of $S$, so by the definition \eqref{eq:hybrid-is-amm} of the hybrid estimator
$$
\phat_\dde^{\mathrm{hyb}}(S)=\widehat M_{E(S)}(S)=\frac{F_1(S)}{n}\,\mathbf 1\{S\in B_n\}.
$$
Let $D_n$ be the uniform distribution on $[n]$ and let $S\sim D_n^n$. Then
$$
\Ex\left[F_1(S)\right]=n\left(1-\frac1n\right)^{n-1}\geq\frac ne
\qquad\text{and}\qquad
\Ex\left[F_2(S)\right]=\binom n2\frac1n\left(1-\frac1n\right)^{n-2}=\frac{n-1}2\left(1-\frac1n\right)^{n-2},
$$
so that $\Ex[F_1(S)-2F_2(S)]=0$. Replacing one sample point changes each of $F_1$ and $F_2$ by at most $2$, hence $F_1-2F_2$ by at most $6$, and McDiarmid's inequality gives $\Pr[S\notin B_n]\leq2e^{-2}$. Since $F_1(S)/n\leq1$,
$$
\Ex\left[\frac{F_1(S)}{n}\mathbf 1\{S\in B_n\}\right]\geq\Ex\left[\frac{F_1(S)}{n}\right]-\Pr[S\notin B_n]\geq\frac1e-\frac2{e^2},
$$
and by the Cauchy--Schwarz inequality, $\Ex[X\mathbf 1\{S\in B_n\}]^2\leq\Ex[X^2\mathbf 1\{S\in B_n\}]$ for $X=F_1(S)/n$, we conclude that
$$
\Ex_{S\sim D_n^n}\left[\left(\phat_\dde^{\mathrm{hyb}}(S)-D_n(\dde(S))\right)^2\right]
=\Ex\left[\frac{F_1(S)^2}{n^2}\mathbf 1\{S\in B_n\}\right]
\geq\left(\frac1e-\frac2{e^2}\right)^2>0
$$
for every $n$. This proves the proposition with $c=(1/e-2/e^2)^2$.

\subsection{A Learner for Which the Hybrid Beats Both Standard Estimators}
\label{sec:hybrid-separation}

The following example shows that holding the base classifier fixed across the folds can be essential. It gives an exception-memorizing learner whose error the hybrid estimates at rate $O(1/n)$ uniformly over all distributions, while both the empirical and the leave-one-out estimator have error bounded away from zero.

Let $\Xcal=\{w\}\cup\Rcal$, where $\Rcal:=\{r_1,r_2,\ldots\}$ is a set of rare inputs. The base learner is unstable in the simplest possible way: it returns a perfect classifier when the sample size is odd and a classifier that errs exactly on $\Rcal$ when the sample size is even, so that $E(S)=\emptyset$ for odd $n$ and $E(S)=\Rcal$ for even $n$. (A symmetric rule that depends on the sample rather than on $n$, such as the parity of the number of distinct sample points, has the same effect.) Concretely, label $y(w)=0$ and $y(r_j)=1$ for every $j$; the learner returns the true labeling at odd sample sizes and the constant classifier $0$ at even ones. The exception-memorizing learner built from it has error region $\dde(S)=E(S)\setminus T(S)$.

\begin{proposition}[The hybrid succeeds where both standard estimators fail]
\label{prop:hybrid-necessary}
For the learner above,
$$
\sup_{D\in\Delta\Xcal}\Ex_{S\sim D^n}\Big[\big(\phat_\dde^{\mathrm{hyb}}(S)-D(\dde(S))\big)^2\Big]=O\Big(\frac1n\Big)
\qquad\text{for every } n,
$$
whereas for every even $n$ there is a distribution $D_n$ at which both the empirical and the leave-one-out estimator have mean-squared error at least $1/16$.
\end{proposition}

\begin{proof}
Since $E(S)$ depends only on $n$, the selection rule $S\mapsto E(S)$ takes a single value, so the first claim follows from Definition~\ref{def:hybrid} and part~\ref{item:amm-info} of Theorem~\ref{thm:adaptive-missing-mass} with $L=1$. For the lower bounds, let $n$ be even and let $D_n$ put mass $1/2$ on $w$ and mass $1/(2n^2)$ on each of $r_1,\ldots,r_{n^2}$. Then $D_n(\dde(S))=\tfrac12-D_n(\Rcal\cap T(S))$, and since each $r_j$ occurs in $S$ with probability at most $n\cdot\frac1{2n^2}=\frac1{2n}$,
$$
\Ex\big[D_n(\Rcal\cap T(S))\big]=\sum_{j=1}^{n^2}\frac1{2n^2}\Pr[r_j\in T(S)]\leq\frac1{4n},
\qquad\text{so that}\qquad
\Ex\big[D_n(\dde(S))\big]\geq\frac12-\frac1{4n}\geq\frac14 .
$$
The empirical estimator is identically zero because $\dde(S)$ contains no sample point. Every leave-one-out fold $S^{-i}$ has odd size, so $\dde(S^{-i})=\emptyset$ and the leave-one-out estimator is identically zero as well. For either estimator, Jensen's inequality gives
\[
\Ex\Big[\big(0-D_n(\dde(S))\big)^2\Big]\geq\big(\Ex[D_n(\dde(S))]\big)^2\geq\frac1{16}. \qedhere
\]
\end{proof}

\subsection{An Information Bound for Adaptive Missing Mass}
\label{app:information-missing-mass}

This section and the next (Appendix~\ref{app:vc-missing-mass}) prove Theorem~\ref{thm:adaptive-missing-mass}. We use the notation of Section~\ref{sec:amm}: for a region $V\subseteq\Xcal$ and a sample $S$,
$$
M_V(S)=D\big(V\setminus T(S)\big),
\qquad
\widehat M_V(S)=\frac1n\sum_{i=1}^n\mathbf 1\{x_i\in V,\ \#_{x_i}(S)=1\},
\qquad
Z_V(S):=\widehat M_V(S)-M_V(S)
$$
are the missing mass of $V$, its restricted Good--Turing estimator, and the error of that estimator. A region selection rule is a map $\Vcal=\Vcal(S,r)$, where $r\sim R$ is an internal seed independent of $S$ (Definition~\ref{def:est-general}); a deterministic rule ignores $r$. For a selection rule we write $Z_\Vcal(S)$ for $Z_{\Vcal(S,r)}(S)$, and $I_D(S;\Vcal)$ for the mutual information, in nats, between $S$ and $\Vcal(S,r)$ when $S\sim D^n$.

\begin{theorem}[Information bound for adaptive missing mass]
\label{thm:information-missing-mass}
There is an absolute constant $c$ such that for every $n$, every $D\in\Delta\Xcal$, and every region selection rule $\Vcal$,
\begin{equation} \label{eq:information-missing-mass}
\Ex\big[Z_\Vcal(S)^2\big]\leq c\,\frac{I_D(S;\Vcal)+1}{n},
\end{equation}
where the expectation is over $S\sim D^n$ and the seed $r\sim R$. In particular, if $\Vcal$ takes at most $L$ values then $\Ex[Z_\Vcal(S)^2]\leq c\,(\log L+1)/n$.
\end{theorem}

The proof has two steps. First, for a region fixed in advance, the error $Z_V(S)$ is sub-Gaussian at scale $1/\sqrt n$, so that $\exp(c'nZ_V(S)^2)$ has bounded expectation for a small absolute constant $c'$. Second, a change of measure from the joint law of $(S,\Vcal)$ to the product of its marginals, at a price of $I_D(S;\Vcal)$, transfers this bound to the selected region.

\begin{proof}
\emph{Step 1: a fixed region.} Fix $D$ and a region $V\subseteq\Xcal$, and write $p_x:=D(x)$. Since $x_i$ is a singleton of $S$ exactly when it does not occur in $S^{-i}$,
$$
\Ex\,\widehat M_V(S)=\Pr\big[x_1\in V,\ x_1\notin T(S^{-1})\big]=\sum_{x\in V}p_x(1-p_x)^{n-1},
\qquad
\Ex\,M_V(S)=\sum_{x\in V}p_x(1-p_x)^{n},
$$
so the bias of the estimator is
\begin{equation} \label{eq:restricted-gt-bias}
0\leq\Ex\,Z_V(S)=\sum_{x\in V}p_x^2(1-p_x)^{n-1}\leq\frac1n\sum_{x\in V}p_x\leq\frac1n,
\end{equation}
using $p(1-p)^{n-1}\leq1/n$ for $p\in[0,1]$.

Both terms of $Z_V$ concentrate. The estimator $\widehat M_V(S)=\frac1n\,|\{x\in V:\#_x(S)=1\}|$ changes by at most $2/n$ when one sample point is replaced, since only the singleton status of the old and of the new value can change, so McDiarmid's inequality gives
\begin{equation} \label{eq:restricted-singleton-concentration}
\Pr\big[|\widehat M_V(S)-\Ex\widehat M_V(S)|\geq t\big]\leq2e^{-nt^2/2}.
\end{equation}
The missing mass $M_V(S)=\sum_{x\in V}p_x\mathbf 1\{\#_x(S)=0\}$ does not have bounded differences, but it concentrates at the same scale. The proofs of \citet[Theorems~10 and~16]{mcallester2003concentration} for the unrestricted missing mass ($V=\Xcal$) use only the negative association of the indicators $\mathbf 1\{\#_x(S)=0\}$, which is inherited by every subcollection, together with bounds on $\sum_{x}p_x^2(1-p_x)^n$ and on $\sum_x p_x^2/\log\big((1-p_x)^{-n}\big)$, which can only decrease when the sums are restricted to $x\in V$. Hence there is an absolute constant $c_1>0$ such that
\begin{equation} \label{eq:restricted-missing-concentration}
\Pr\big[|M_V(S)-\Ex M_V(S)|\geq t\big]\leq2e^{-c_1nt^2}.
\end{equation}
Let $X_V:=Z_V(S)-\Ex Z_V(S)$. Since $|X_V|\geq t$ requires one of the two centered terms to deviate by at least $t/2$, \eqref{eq:restricted-singleton-concentration} and \eqref{eq:restricted-missing-concentration} give $\Pr[|X_V|\geq t]\leq4e^{-c_2nt^2}$ with $c_2:=\min\{1/8,\,c_1/4\}$. Integrating this tail with $\lambda:=c_2n/2$,
$$
\Ex\,e^{\lambda X_V^2}
= 1+\int_0^\infty\lambda e^{\lambda u}\Pr\big[X_V^2>u\big]\,du
\leq1+4\lambda\int_0^\infty e^{-(c_2n-\lambda)u}\,du
=5.
$$
Finally $Z_V^2\leq2X_V^2+2(\Ex Z_V)^2\leq2X_V^2+2/n^2$ by \eqref{eq:restricted-gt-bias}, so with $c_3:=c_2/4$ we obtain, for every region $V$, every $D$ \new{and every $n\geq1$ (since $c_2\leq1/8$, so that $5e^{c_2/(2n)}\leq5e^{1/16}<6$)},
\begin{equation} \label{eq:restricted-gt-exponential-square}
\Ex_{S\sim D^n}\exp\big(c_3nZ_V(S)^2\big)\leq5e^{c_2/(2n)}\leq6.
\end{equation}

\emph{Step 2: decoupling.} Let $P$ be the joint law of $(S,\Vcal)$ with $S\sim D^n$ and $\Vcal=\Vcal(S,r)$, and let $Q:=P_S\otimes P_\Vcal$ be the product of its marginals, so that $D_{\mathrm{KL}}(P\,\|\,Q)=I_D(S;\Vcal)$. If $I_D(S;\Vcal)=\infty$ there is nothing to prove. Otherwise the Donsker--Varadhan variational inequality, $\Ex_P[f]\leq D_{\mathrm{KL}}(P\,\|\,Q)+\log\Ex_Q[e^f]$, applied to $f(S,\Vcal):=c_3nZ_\Vcal(S)^2$ gives
$$
c_3n\,\Ex_P\big[Z_\Vcal(S)^2\big]
\leq I_D(S;\Vcal)+\log\Ex_Q\exp\big(c_3nZ_\Vcal(S)^2\big)
\leq I_D(S;\Vcal)+\log6.
$$
The last step uses that under $Q$ the region $\Vcal$ is independent of $S$, so \eqref{eq:restricted-gt-exponential-square} applies conditionally on $\Vcal$ and hence after averaging over its law. Rearranging proves \eqref{eq:information-missing-mass} with $c:=(\log6)/c_3$. If $\Vcal$ takes at most $L$ values then $I_D(S;\Vcal)\leq H(\Vcal)\leq\log L$.
\end{proof}

\subsection{VC Bounds for Adaptive Missing Mass}
\label{app:vc-missing-mass}

Theorem~\ref{thm:information-missing-mass} gives the rate $(\log L+1)/n$ for a selection rule with at most $L$ possible outputs. We now replace $\log L$ by the VC dimension of the class of possible outputs, with no restriction on how the region is chosen from that class. Throughout, $\Vcal=\Vcal(S,r)$ is a region selection rule with values in a class $\Hcal\subseteq2^{\Xcal}$, $Z_\Vcal(S)=\widehat M_{\Vcal(S)}(S)-M_{\Vcal(S)}(S)$ is its error as in Appendix~\ref{app:information-missing-mass}, and $\hat D$ is the empirical distribution of $S$ (Section~\ref{sec:empest}).
\new{As in Appendix~\ref{app:defvc}, $\|\hat D-D\|_{\Fcal}$ denotes the uniform deviation of the empirical measure over a class $\Fcal$, and we write} $\Hcal\,\Delta\,\Hcal:=\{V\,\Delta\,V':V,V'\in\Hcal\}$ for the class of symmetric differences of pairs of regions in $\Hcal$.

The argument is a covering reduction, applied to one distribution $D$ at a time \new{(the cover is used only in the analysis, so it may depend on $D$)}. For $V,V'\subseteq\Xcal$ let $\rho_D(V,V'):=D(V\,\Delta\,V')$. An \emph{internal $\eta$-cover} of $\Hcal$ is a finite $\Bcal\subseteq\Hcal$ such that every $V\in\Hcal$ has some $B\in\Bcal$ with $\rho_D(V,B)\leq\eta$; let $\Ncal_D(\eta,\Hcal)$ denote the smallest size of such a cover. We replace the selected region by a nearby element of a cover, apply Theorem~\ref{thm:information-missing-mass} to the resulting finite-valued rule, and control the effect of the replacement by uniform convergence over $\Hcal\,\Delta\,\Hcal$. Requiring the cover to lie inside $\Hcal$ restricts the replacement errors to this class. \new{The two lemmas below and Lemma~\ref{lem:amm-uniform} supply the ingredients.}

\begin{lemma}[Replacing the region] \label{lem:amm-projection}
For all regions $V,V'\subseteq\Xcal$ and every sample $S$,
$$
|Z_V(S)-Z_{V'}(S)|\leq\hat D(V\,\Delta\,V')+D(V\,\Delta\,V').
$$
\end{lemma}

\begin{proof}
A sample point contributes to exactly one of $\widehat M_V(S)$ and $\widehat M_{V'}(S)$ only if it lies in $V\,\Delta\,V'$, so $|\widehat M_V(S)-\widehat M_{V'}(S)|\leq\frac1n\sum_{i=1}^n\mathbf 1\{x_i\in V\,\Delta\,V'\}=\hat D(V\,\Delta\,V')$. Also $|M_V(S)-M_{V'}(S)|=|D(V\setminus T(S))-D(V'\setminus T(S))|\leq D(V\,\Delta\,V')$.
\end{proof}

\begin{lemma}[Symmetric differences] \label{lem:amm-vc-difference}
If $\operatorname{VC}(\Hcal)=d$ then $\operatorname{VC}(\Hcal\,\Delta\,\Hcal)\leq10d$.
\end{lemma}

\begin{proof}
If $d=0$ then $\Hcal$ has at most one member and $\Hcal\,\Delta\,\Hcal\subseteq\{\emptyset\}$. Let $d\geq1$ and let $\Pi_\Hcal(m)$ be the growth function of $\Hcal$, the largest number of distinct traces of $\Hcal$ on $m$ points. Every trace of $\Hcal\,\Delta\,\Hcal$ on a set of $m$ points is the symmetric difference of two traces of $\Hcal$, so $\Pi_{\Hcal\Delta\Hcal}(m)\leq\Pi_\Hcal(m)^2$. If $\Hcal\,\Delta\,\Hcal$ shattered $m=10d$ points, Sauer's lemma would give $2^{10d}\leq\Pi_\Hcal(10d)^2\leq(10e)^{2d}$, which is false since $(10e)^2<740<1024=2^{10}$.
\end{proof}

\begin{theorem}[Covering bound for adaptive missing mass] \label{thm:amm-cover}
Let $\operatorname{VC}(\Hcal)=d<\infty$ and let $\Vcal$ be a selection rule with values in $\Hcal$. There is an absolute constant $c$ such that for every $D\in\Delta\Xcal$ and every $0<\eta\leq1$,
\begin{equation} \label{eq:amm-cover-explicit}
\Ex\big[Z_\Vcal(S)^2\big]
\leq c\left(\eta^2+\frac{\log\Ncal_D(\eta,\Hcal)+1}{n}+\frac{d+1}{n}\right).
\end{equation}
\end{theorem}

\begin{proof}
Fix $D$ and $\eta$; if $\Ncal_D(\eta,\Hcal)=\infty$ there is nothing to prove. Fix an internal $\eta$-cover $\Bcal=\{B_1,\ldots,B_L\}\subseteq\Hcal$ of size $L=\Ncal_D(\eta,\Hcal)$, and let $\pi:\Hcal\to\Bcal$ map each region to the first cover element within $\rho_D$-distance $\eta$ of it. Then $B:=\pi(\Vcal)$ is a selection rule with at most $L$ values, so Theorem~\ref{thm:information-missing-mass} gives $\Ex[Z_B(S)^2]\leq c_0(\log L+1)/n$ for an absolute constant $c_0$. Since $\Vcal(S)\,\Delta\,B(S)\in\Hcal\,\Delta\,\Hcal$ and $D(\Vcal(S)\,\Delta\,B(S))\leq\eta$, Lemma~\ref{lem:amm-projection} gives
$$
|Z_\Vcal(S)-Z_B(S)|\leq\hat D(\Vcal(S)\,\Delta\,B(S))+\eta\leq2\eta+\|\hat D-D\|_{\Hcal\Delta\Hcal}.
$$
Therefore
$$
\Ex\big[Z_\Vcal(S)^2\big]
\leq2\Ex\big[Z_B(S)^2\big]+2\Ex\big[(Z_\Vcal(S)-Z_B(S))^2\big]
\leq\frac{2c_0(\log L+1)}{n}+16\eta^2+4\Ex\big[\|\hat D-D\|_{\Hcal\Delta\Hcal}^2\big],
$$
and by Lemmas~\ref{lem:amm-vc-difference} and~\ref{lem:amm-uniform} the last expectation is $O((d+1)/n)$.
\end{proof}

Haussler's packing bound \citep{haussler1995sphere} controls the covering numbers uniformly in $D$: for every class of VC dimension $d\geq1$ and every $0<\eta\leq1$,
\begin{equation} \label{eq:amm-haussler}
\log\Ncal_D(\eta,\Hcal)\leq c_1\,d\log\frac{c_2}{\eta}
\end{equation}
for absolute constants $c_1,c_2\geq e$. Choosing $\eta$ to balance the terms of \eqref{eq:amm-cover-explicit} gives part~\ref{item:amm-vc} of Theorem~\ref{thm:adaptive-missing-mass}.

\begin{corollary}[VC bound for adaptive missing mass] \label{cor:vc-missing-mass}
Let $\Vcal$ be a selection rule with values in a class $\Hcal$ with $\operatorname{VC}(\Hcal)=d$. If $1\leq d\leq n$ then
\begin{equation} \label{eq:amm-vc-final}
\sup_{D\in\Delta\Xcal}\Ex\big[Z_\Vcal(S)^2\big]
=O\Big(\frac{d\log(en/d)}{n}\Big)
\end{equation}
with an absolute constant, and if $d=0$ the left-hand side is $O(1/n)$. In particular, if $d_n:=\operatorname{VC}(\Hcal_n)=o(n)$ then the left-hand side tends to $0$.
\end{corollary}

\begin{proof}
For $1\leq d\leq n$ apply Theorem~\ref{thm:amm-cover} with $\eta=\sqrt{d/n}\leq1$ and use \eqref{eq:amm-haussler}: since $\log(en/d)\geq1$, each of the three terms in \eqref{eq:amm-cover-explicit} is $O(d\log(en/d)/n)$. If $d=0$ then $\Hcal$ has a single member, and Theorem~\ref{thm:information-missing-mass} with $L=1$ gives $O(1/n)$. For the last claim, $x\log(e/x)\to0$ as $x=d_n/n\to0$.
\end{proof}

\subsection{Proof of Corollary~\ref{cor:vc-hypothesis-class} and the Multiclass Extension}
\label{app:vc-hypothesis-class}

Throughout, $\Xcal_y$, $\Ecal_n^y$, and $\dde_y$ are defined as in Section~\ref{sec:beyond} for a labeling $y:\Wcal\to\Ycal$.

\begin{proof}[Proof of Corollary~\ref{cor:vc-hypothesis-class}]
\emph{Error regions.} Fix a labeling $y$ and, for $h\in\Hcal$, let $E_h:=\{w\in\Wcal:h(w)\neq y(w)\}$ be its error region relative to $y$, whose indicator is $h\oplus y$. For a finite $T\subseteq\Wcal$, the map $b\mapsto b\oplus y|_T$ is a bijection of $\zo^T$, and it maps the set of restrictions to $T$ of the members of $\Hcal$ onto the set of restrictions to $T$ of the members of $\{E_h:h\in\Hcal\}$. Hence $T$ is shattered by $\{E_h:h\in\Hcal\}$ if and only if it is shattered by $\Hcal$, and $\operatorname{VC}(\{E_h:h\in\Hcal\})=d$. Since the output of $\Acal$ lies in $\Hcal$, \new{every member of $\Ecal_n^y$ is the image of some $E_h$ under the bijection $w\mapsto(w,y(w))$ from $\Wcal$ to $\Xcal_y$, which preserves the VC dimension}, and so $\operatorname{VC}(\Ecal_n^y)\leq d$.

\emph{The estimate.} Fix $y$. By Definition~\ref{def:hybrid}, the error of the hybrid estimator for $\dde_y$ is $\varepsilon^{\mathrm{amm}}_{E^y}(n)$ for the selection rule $S\mapsto E^y(S)$ on the example space $\Xcal_y$, whose image class is $\Ecal_n^y$. Apply Theorem~\ref{thm:adaptive-missing-mass} with $d_n:=\operatorname{VC}(\Ecal_n^y)\leq d\leq n$. If $d_n\geq1$, part~\ref{item:amm-vc} gives $O(d_n\log(en/d_n)/n)$, which is $O(d\log(en/d)/n)$ because $x\mapsto x\log(en/x)$ is nondecreasing on $(0,n]$. If $d_n=0$ then $\Ecal_n^y$ consists of a single region, and part~\ref{item:amm-info} with $L=1$ gives $O(1/n)$, which is again $O(d\log(en/d)/n)$ as $d\geq1$. The constants in Theorem~\ref{thm:adaptive-missing-mass} are absolute and do not depend on the example space, so the bound holds with one constant for every $y$.
\end{proof}

For a general label set the role of the VC dimension is played by the graph dimension. For a class $\Hcal\subseteq\Ycal^{\Wcal}$, write $\Hcal^{\mathrm{gr}}:=\{(w,y')\mapsto\mathbf 1\{h(w)=y'\}:h\in\Hcal\}$ for the class of graphs of its members, viewed as subsets of $\Wcal\times\Ycal$; the \emph{graph dimension} of $\Hcal$ is $\operatorname{VC}(\Hcal^{\mathrm{gr}})$ \citep{natarajan1989learning,bendavid1995characterizations}. When $\Ycal=\zo$ it coincides with $\operatorname{VC}(\Hcal)$. Indeed, if $T\subseteq\Wcal$ is shattered by $\Hcal$ then $\{(w,1):w\in T\}$ is shattered by $\Hcal^{\mathrm{gr}}$, since $\mathbf 1\{h(w)=1\}=h(w)$. Conversely, let $U\subseteq\Wcal\times\zo$ be shattered by $\Hcal^{\mathrm{gr}}$. No $h$ satisfies $h(w)=0$ and $h(w)=1$ at once, so $U$ contains at most one pair with each first coordinate, and $U=\{(w,y'_w):w\in T\}$ for some $T\subseteq\Wcal$ with $|T|=|U|$. For every $b\in\zo^T$ there is $h\in\Hcal$ with $h(w)=y'_w$ when $b(w)=1$ and $h(w)=1-y'_w$ when $b(w)=0$, and as $b$ ranges over $\zo^T$ so does this labeling of $T$; thus $T$ is shattered by $\Hcal$. Corollary~\ref{cor:vc-hypothesis-class} is therefore the case $\Ycal=\zo$ of the following statement.

\begin{corollary}[Multiclass version of Corollary~\ref{cor:vc-hypothesis-class}] \label{cor:vc-hypothesis-class-multi}
Let $\Ycal$ be any label set, let $\Hcal\subseteq\Ycal^{\Wcal}$ be a class with graph dimension $d$, and let $\Acal$ be a base learner whose output lies in $\Hcal$ on every labeled sample. Then $\operatorname{VC}(\Ecal_n^y)\leq d$ for every labeling $y:\Wcal\to\Ycal$ and every $n$, and if $1\leq d\leq n$ then
$$
\sup_{D\in\Delta\Xcal_y}\Ex_{S\sim D^n}\Big[\big(\phat_{\dde_y}^{\mathrm{hyb}}(S)-D(\dde_y(S))\big)^2\Big]
=O\Big(\frac{d\log(en/d)}{n}\Big)
$$
for every labeling $y$, with an absolute constant.
\end{corollary}

\begin{proof}
Fix a labeling $y$ and, for $h\in\Hcal$, let $E_h:=\{w\in\Wcal:h(w)\neq y(w)\}$. Suppose that $T\subseteq\Wcal$ is shattered by $\{E_h:h\in\Hcal\}$ and let $U:=\{(w,y(w)):w\in T\}$, so that $|U|=|T|$. For every $b\in\zo^T$ there is $h\in\Hcal$ with $\mathbf 1\{h(w)\neq y(w)\}=b(w)$ for all $w\in T$, that is, $\mathbf 1\{h(w)=y(w)\}=1-b(w)$; hence $U$ is shattered by $\Hcal^{\mathrm{gr}}$ and $|T|\leq d$. \new{Since every member of $\Ecal_n^y$ is the image of some $E_h$ under the bijection $w\mapsto(w,y(w))$}, we get $\operatorname{VC}(\Ecal_n^y)\leq d$. The remaining step of the proof of Corollary~\ref{cor:vc-hypothesis-class} applies verbatim, with $\Ycal$ in place of $\zo$.
\end{proof}

\ifarxiv
\begin{remark}[The logarithmic factor] \label{rem:amm-log}
For a finite class the direct bound $O((\log|\Hcal|+1)/n)$ of Theorem~\ref{thm:information-missing-mass} is preferable to \eqref{eq:amm-vc-final}. For an infinite VC class the factor $\log(en/d)$ comes from applying the finite-range bound to a single cover at one scale; the argument does not show that it is intrinsic, and removing it would require a multiscale or localized analysis of the process $V\mapsto Z_V(S)$.
\end{remark}
\fi
\section{Proof of the Quadratic-Domain Construction}
\label{app:lower-bound-proofs}
\new{We assume $n\geq 2$ throughout.}

\begin{proof}[Proof of Theorem~\ref{claim:quadratic-domain-lb}]
\new{\emph{The construction.} Let $w=n^2$, $m=2w=2n^2$ and $\Xcal_n=[m]$. For a set $T\subseteq[m]$ with $|T|=n$ and $V$ uniform on $\binom{[m]}{w}$, let
\[
 p_n:=\Pr[T\subseteq V]=\frac{\binom{m-n}{w-n}}{\binom{m}{w}}
 =\prod_{\ell=0}^{n-1}\frac{w-\ell}{m-\ell},
\]
which does not depend on $T$, and set $M:=\lfloor 1/p_n\rfloor+1$. Since each factor $(w-\ell)/(m-\ell)$ is at most $1/2$, we have $p_n\leq 2^{-n}$. Draw $V_1,\ldots,V_M$ independently and uniformly from $\binom{[m]}{w}$, write $\mathcal C=(V_1,\ldots,V_M)$, and let $D_i$ be the uniform distribution on $V_i$. For a sample $S=(x_1,\ldots,x_n)\in[m]^n$ let $T(S)=\{x_1,\ldots,x_n\}$, let $I_{\mathcal C}(S)=\{i\in[M]:T(S)\subseteq V_i\}$ be the set of codewords consistent with $S$, and define $\dde_{\mathcal C}(S)=V_{\min I_{\mathcal C}(S)}$ if $I_{\mathcal C}(S)\neq\emptyset$ and $\dde_{\mathcal C}(S)=\emptyset$ otherwise.

\emph{Reduction to a Bayesian experiment.} Fix a code $\mathcal C$ and an estimator $\phat:\Xcal_n^n\to[0,1]$. Let $I$ be uniform on $[M]$ and, given $I$, let $S\sim D_I^n$. Since the supremum over $D\in\Delta(\Xcal_n)$ dominates the average over $D_1,\ldots,D_M$,
\begin{equation}\label{eq:lb-bayes}
\sup_{D\in\Delta(\Xcal_n)}\Ex_{S\sim D^n}\left[\left(\phat(S)-D(\dde_{\mathcal C}(S))\right)^2\right]
\geq
\Ex\left[\left(\phat(S)-D_I(\dde_{\mathcal C}(S))\right)^2\;\middle|\;\mathcal C\right].
\end{equation}
Each $D_i^n$ assigns probability $w^{-n}$ to every ordered sample $S$ with $T(S)\subseteq V_i$ and probability $0$ to every other sample. Hence, given $\mathcal C$ and $S$, the posterior distribution of $I$ is uniform on $I_{\mathcal C}(S)$, which contains $I$. On the event $G:=\{|T(S)|=n,\ |I_{\mathcal C}(S)|=2\}$, write $I_{\mathcal C}(S)=\{a,b\}$, let $J=\min\{a,b\}$, so that $\dde_{\mathcal C}(S)=V_J$, and let $\rho:=|V_a\cap V_b|/w$. The target $D_I(V_J)$ equals $1$ if $I=J$ and $\rho$ otherwise, each with posterior probability $1/2$. Since $\phat(S)$ is determined by $S$, its conditional mean-squared error given $(\mathcal C,S)$ is at least the conditional variance $(1-\rho)^2/4$ of the target. So for every estimator the right-hand side of \eqref{eq:lb-bayes} is at least $B(\mathcal C):=\frac14\Ex\left[(1-\rho)^2\mathbf 1_G\;\middle|\;\mathcal C\right]$, and it suffices to show that $\Ex_{\mathcal C}B(\mathcal C)\geq1/576$: some code $\mathcal C^*$ then has $B(\mathcal C^*)\geq 1/576$, and $\dde_n:=\dde_{\mathcal C^*}$ has the claimed property.

\emph{Lower bound on $\Ex_{\mathcal C}B(\mathcal C)$.} Condition on $I=i$, on $V_i$, and on the sample $S$. Under this conditioning the other codewords $V_j$, $j\neq i$, are still independent and uniform on $\binom{[m]}{w}$. Let $E_0$ be the event that $x_1,\ldots,x_n$ are distinct. Since the sample is drawn uniformly from a set of size $w=n^2$,
\[
\Pr[E_0\mid I,V_I]=\prod_{t=0}^{n-1}\Big(1-\frac tw\Big)\geq 1-\frac{n(n-1)}{2w}\geq\frac12 .
\]
Assume that $E_0$ holds, so that $T:=T(S)$ has exactly $n$ elements. Then $W:=|I_{\mathcal C}(S)|-1=\sum_{j\neq i}\mathbf 1\{T\subseteq V_j\}$ is $\operatorname{Binomial}(M-1,p_n)$ with mean $\lambda:=(M-1)p_n=\lfloor 1/p_n\rfloor p_n\in[1-p_n,1]$, and
\[
\Pr[W=1\mid I,V_I,S]=\lambda(1-p_n)^{M-2}\geq \frac12\cdot\frac14=\frac18,
\]
using $\lambda\geq 1-p_n\geq 1/2$, $M-2\leq 1/p_n$, and $(1-p)^{1/p}\geq 1/4$ for $0<p\leq 1/2$. Given in addition that $W=1$ and that $b$ is the index of the other consistent codeword, the codeword $V_b$ is uniform among the $w$-subsets of $[m]$ that contain $T$, independently of $V_i$: the event depends on $V_b$ only through $T\subseteq V_b$, and on the other codewords only through their not containing $T$. So $V_b\setminus T$ is a uniform $(w-n)$-subset of $[m]\setminus T$, and $|V_i\cap V_b|=n+H$ with $H\sim\operatorname{Hypergeometric}(m-n,\,w-n,\,w-n)$ (a population of $m-n$ elements, of which the $w-n$ elements of $V_i\setminus T$ are marked, and $w-n$ draws). Since $m=2w$ and $w=n^2$,
\[
\Ex\left[1-\rho\;\middle|\;I,V_I,S,W=1,b\right]=1-\frac nw-\frac{(w-n)^2}{w(m-n)}=\frac{w-n}{2w-n}=\frac{n-1}{2n-1}\geq\frac13,
\]
and by Jensen's inequality $\Ex\left[(1-\rho)^2\;\middle|\;I,V_I,S,W=1,b\right]\geq 1/9$. Averaging over $b$, $S$, $V_I$ and $I$, and using that $G\supseteq E_0\cap\{W=1\}$,
\[
\Ex_{\mathcal C}B(\mathcal C)\geq\frac14\cdot\Pr[E_0]\cdot\frac18\cdot\frac19\geq\frac14\cdot\frac12\cdot\frac18\cdot\frac19=\frac1{576}. \qedhere
\]}
\end{proof}

\section{Estimation of Data-dependent Mass versus Its Expectation}
\label{app:classnequnclass}

The value $D(\dde(S))$ in Definition~\ref{def:estmse} is a random quantity, determined by the realized sample. One could instead ask to estimate its expectation,
\begin{equation} \label{eq:expected-mass}
\theta_\dde(D) := \Ex_{S \sim D^n}\big[D(\dde(S))\big],
\end{equation}
a deterministic functional of $D$. This is the form of problem to which the classical lower-bound techniques for functional estimation, the two-point method of Le~Cam and the method of Fano \citep{lecam1973convergence,fano1966transmission,yu1997assouad}, apply, and one might hope to obtain lower bounds on estimatability from them. The following example shows that the two targets are different: the random mass can be observed exactly while its expectation cannot be estimated to within a constant. A lower bound on estimating \eqref{eq:expected-mass} therefore carries no implication for estimatability in the sense of Definition~\ref{def:estmse}, which is why the impossibility results of Section~\ref{sec:limits} argue directly about the conditional distribution of the random target given the sample.

\begin{proposition} \label{app:thmsplit}
Let $\Xcal = \mathbb{N} \cup \{0\}$ and let $\dde(S) = \Xcal$ if $0 \in S$ and $\dde(S) = \emptyset$ otherwise. Then the estimator $\phat(S) = \mathbf 1\{0 \in S\}$ satisfies $\phat(S) = D(\dde(S))$ for every sample and every $D$, so $\dde$ is estimatable with zero error, whereas for every $n \geq 1$
$$
\inf_{\phat} \sup_{D \in \Delta \Xcal} \Ex_{S \sim D^n}\Big[\big(\phat(S) - \theta_\dde(D)\big)^2\Big] \geq \new{\frac{(1-e^{-1/2})^2}{16}}.
$$
\end{proposition}

\begin{proof}
The first claim holds because $D(\dde(S)) = \mathbf 1\{0 \in S\}$ is a function of the sample.

For the second claim, note that $\theta_\dde(D)$ is the probability that the point $0$ appears in the sample,
$$
\theta_\dde(D) = \Pr_{S \sim D^n}[0 \in S] = 1 - (1 - D(0))^n .
$$
We compare two distributions that differ only in the mass of the point $0$. Let $\delta := 1/(2n)$, let $D_0$ put all of its mass on the point $1$, and let $D_1$ put mass $\delta$ on $0$ and mass $1 - \delta$ on $1$. Under $D_0$ the point $0$ never appears, so $\theta_\dde(D_0) = 0$, whereas
$$
\theta_\dde(D_1) = 1 - (1-\delta)^n \geq 1 - e^{-n\delta} = 1 - e^{-1/2} .
$$
Let $\Delta := \theta_\dde(D_1) - \theta_\dde(D_0)$ denote the gap between the two values, so $\Delta \geq 1 - e^{-1/2}$. The two sample laws, on the other hand, are close: $D_0$ and $D_1$ differ by $\delta$ in total variation, so
$$
\operatorname{TV}(D_0^n, D_1^n) \leq n\,\operatorname{TV}(D_0, D_1) = n\delta = \frac12 .
$$
\new{By the two-point method of Le~Cam \citep{lecam1973convergence,yu1997assouad}, every estimator $\phat$ therefore satisfies
$$
\max_{j\in\{0,1\}}\Ex_{S \sim D_j^n}\Big[\big(\phat(S)-\theta_\dde(D_j)\big)^2\Big]\geq\frac{\Delta^2}{8}\big(1-\operatorname{TV}(D_0^n,D_1^n)\big)\geq\frac{(1-e^{-1/2})^2}{16}.
$$}
\end{proof}

\bibliographystyle{plainnat}

\end{document}